\documentclass[letterpaper,10pt]{IEEEtran}
\usepackage{amsmath,amssymb,amsthm,mathtools,bm}
\usepackage{algorithm}
\usepackage{algorithmicx}
\usepackage[noEnd=false,indLines=true]{algpseudocodex}

\algrenewcommand\algorithmicrequire{\textbf{Input:}}
\algrenewcommand\algorithmicensure{\textbf{Output:}}

\usepackage{array,booktabs,multirow,adjustbox}
\usepackage[caption=false,font=footnotesize]{subfig}
\usepackage{graphicx}
\usepackage{cite}
\usepackage{stfloats}
\usepackage{placeins}
\usepackage{float}
\usepackage[table]{xcolor}
\usepackage{tikz}
\usepackage{microtype}
\usetikzlibrary{arrows.meta,positioning,calc,fit,shapes.geometric,shapes.misc}
\usepackage{url}
\usepackage{balance}
\usepackage[percent]{overpic}
\usepackage{tikz}
\usetikzlibrary{calc}

\definecolor{easybg}{RGB}{226,239,218}
\definecolor{mediumbg}{RGB}{255,249,207}
\definecolor{hardbg}{RGB}{250,222,220}
\definecolor{overallbg}{RGB}{236,236,236}
\definecolor{stageone}{RGB}{244,238,252}
\definecolor{stagetwo}{RGB}{235,244,253}
\definecolor{stagethree}{RGB}{238,248,241}
\definecolor{lineblue}{RGB}{53,91,141}
\definecolor{linegreen}{RGB}{65,115,82}
\definecolor{mocknum}{RGB}{92,103,115}
\definecolor{lightgrid}{RGB}{220,224,229}
\definecolor{targetyellow}{RGB}{248,209,65}
\newcounter{cportheorem}
\newcounter{cporcorollary}
\newcommand{\theoremhead}[2]{%
  \refstepcounter{cportheorem}\label{#1}%
  \noindent\textbf{Theorem \thecportheorem\ (#2).}\ }
\newcommand{\corollaryhead}[2]{%
  \refstepcounter{cporcorollary}\label{#1}%
  \noindent\textbf{Corollary \thecporcorollary\ (#2).}\ }

\newcommand{\TBD}{\textcolor{mocknum}{--}}

\newcommand{\bfe}{\mathbf e}
\newcommand{\cD}{\mathsf{D}}
\newcommand{\Reach}{\operatorname{Reach}}
\newcommand{\Free}{\operatorname{Free}}
\newcommand{\logit}{\operatorname{logit}}

\DeclareMathOperator*{\argmin}{arg\,min}
\newcommand{\clip}{\operatorname{clip}}
\newcommand{\cP}{\mathcal{P}}
\newcommand{\cR}{\mathcal{R}}
\newcommand{\cA}{\mathcal{A}}
\newcommand{\cC}{\mathcal{C}}
\newcommand{\indic}{\mathbb{I}}
\newcommand{\PD}{P_{\mathcal D}}
\newcommand{\PK}{P_K}
\newcommand{\Lse}{\operatorname{LSE}}
\newcommand{\dTV}{d_{\mathrm{TV}}}

\theoremstyle{definition}

\usepackage{pgfplots}
\usepgfplotslibrary{groupplots}
\usetikzlibrary{arrows.meta,positioning,calc}
\usepackage[hidelinks]{hyperref}
\pgfplotsset{compat=1.18}
\usepackage{orcidlink}

\renewcommand{\arraystretch}{1.13}

\definecolor{EasyColor}{RGB}{221,235,209}
\definecolor{MediumColor}{RGB}{255,248,196}
\definecolor{HardColor}{RGB}{250,211,211}
\definecolor{OverallColor}{RGB}{235,235,235}
\definecolor{CoreColor}{RGB}{230,238,249}
\definecolor{AuditColor}{RGB}{239,234,249}
\definecolor{RobotColor}{RGB}{231,242,232}
\definecolor{WarnColor}{RGB}{252,236,230}
\definecolor{MockBlue}{RGB}{43,93,158}
\definecolor{MockGray}{RGB}{248,249,251}
\definecolor{GoodColor}{RGB}{228,242,231}
\definecolor{CautionColor}{RGB}{255,246,218}
\definecolor{ProposedColor}{RGB}{76,111,163}
\definecolor{BaselineColor}{RGB}{115,115,115}
\definecolor{FusionColor}{RGB}{79,135,190}
\definecolor{DAGColor}{RGB}{96,72,153}
\definecolor{DeferColor}{RGB}{79,133,95}
\definecolor{easybg}{RGB}{220,235,211}
\definecolor{mediumbg}{RGB}{255,249,198}
\definecolor{hardbg}{RGB}{249,213,213}
\definecolor{overallbg}{RGB}{232,232,232}

\DeclareRobustCommand{\targetstar}{%
  \tikz[baseline=-0.65ex]\node[
    star, star points=5, star point ratio=2.25,
    minimum size=1.55ex, inner sep=0pt,
    fill=yellow, draw=black, line width=0.45pt
  ] {};%
}

\DeclareRobustCommand{\topobstructor}{%
  \tikz[baseline=-0.60ex]\node[
    rectangle, minimum size=1.35ex, inner sep=0pt,
    fill=yellow, draw=black, line width=0.45pt
  ] {};%
}

\DeclareRobustCommand{\unograspmarker}{%
  \tikz[baseline=-0.58ex]\node[
    circle, minimum size=1.35ex, inner sep=0pt,
    fill=red!85, draw=black, line width=0.45pt
  ] {};%
}

\DeclareRobustCommand{\bluediamond}{%
  \tikz[baseline=-0.62ex]\node[
    diamond, aspect=1.15,
    minimum size=1.50ex, inner sep=0pt,
    fill=cyan!45, draw=blue!75!black, line width=0.50pt
  ] {};%
}

\DeclareRobustCommand{\greenroundbox}{%
  \tikz[baseline=-0.57ex]\node[
    rounded rectangle, rounded rectangle arc length=120,
    minimum width=1.65ex, minimum height=1.18ex, inner sep=0pt,
    fill=green!55, draw=green!30!black, line width=0.50pt
  ] {};%
}

\title{%CPOR-Grasp: \textbf{C}alibrated \textbf{P}robabilistic \textbf{O}bstruction \textbf{R}easoning with Vision-Language Models for Grasping in Clutter}
Calibrated Probabilistic Obstruction Reasoning with Vision-Language Models for Grasping in Clutter}
\author{Thanh-Tuan Tran\orcidlink{https://orcid.org/0009-0006-4370-8326}, Ngoc-Chien Chu, Thanh Nguyen Canh\orcidlink{https://orcid.org/0000-0001-6332-1002}~\IEEEmembership{Graduate Student Member,~IEEE}, Nak Young Chong\orcidlink{https://orcid.org/0000-0001-5736-0769}~\IEEEmembership{Senior Member,~IEEE}, Nguyen Viet Ha and Xiem HoangVan\orcidlink{https://orcid.org/0000-0001-6332-1002} % <-this % stops a space
\thanks{This work was supported in part by JST SPRING, Japan Grant Number JPMJSP2102 , and in part by the Asian Office of Aerospace Research and Development under Grant/Cooperative Agreement Award No. FA2386-25-1-4034. (Corresponding author: Xiem HoangVan)}
\thanks{Thanh Tuan Tran, Ngoc Chien Chu, Nguyen Viet Ha and Xiem HoangVan is with the University of Engineering and Technology, Vietnam National University, Hanoi, 10000, Vietnam.  {\footnotesize \{, , , xiemhoang\}@vnu.edu.vn}}%
\thanks{Thanh Nguyen Canh and Nak Yong Chong are with the School of Information Science, Japan Advanced Institute of Science and Technology, Ishikawa, 923-1211, Japan. {\footnotesize \{thanhnc, nakyoung\}@jaist.ac.jp}; Nak Yong Chong is also with the Department of Robotics, Hanyang University, Gyeonggi-do, 15588, Korea. {\footnotesize nychong@hanyang.ac.kr}}%

\thanks{Code and video are publicly available at \href{https://anonymous.4open.science/r/cpor-grasp-1E41}{\textcolor{blue}{Cpor-grasp}}.}
}

\begin{document}
\setlength{\parindent}{1em}
\setlength{\parskip}{0pt}
\maketitle

\begin{abstract}
Retrieving a target from clutter requires deciding whether to grasp the target, remove a blocker, or defer. Existing methods typically commit to a single obstruction graph or removal strategy, ignoring uncertainty across alternative scene interpretations. They also rely on miscalibrated vision-language model (VLM) predictions and can produce pairwise obstruction relations that are jointly inconsistent. Moreover, current approximations provide no guarantees about the impact of discarded hypotheses on the final decision.
We propose CPOR-Grasp, a calibrated probabilistic obstruction-reasoning framework that propagates uncertainty from pairwise evidence to action decisions. CPOR-Grasp calibrates and fuses VLM, depth, and amodal-mask cues to estimate obstruction probabilities, induces a distribution over valid obstruction graphs, and marginalizes over these graphs to compute the likelihood that the target is accessible or that a given blocker should be removed. To make inference tractable, it retains only the highest-probability graphs and derives a total-variation bound on the discarded probability mass, enabling certified decisions, adaptive stopping, and principled deferral.
On synthetic and real UNOBench scenes, CPOR-Grasp outperforms state-of-the-art baselines. Calibration error decreases from 0.1416 to 0.0185 on the Gemini Robotics backbone, while graph truncation matches exact inference on 99.74\% of decisions using 56 times fewer graphs. In real-world experiments, CPOR-Grasp achieves a 77.8\% average success rate, surpassing SOTA baselines.
\end{abstract}


\section{Introduction}
\label{sec:intro}
\begin{figure}[!t]
\centering
\includegraphics[width=0.5\textwidth]{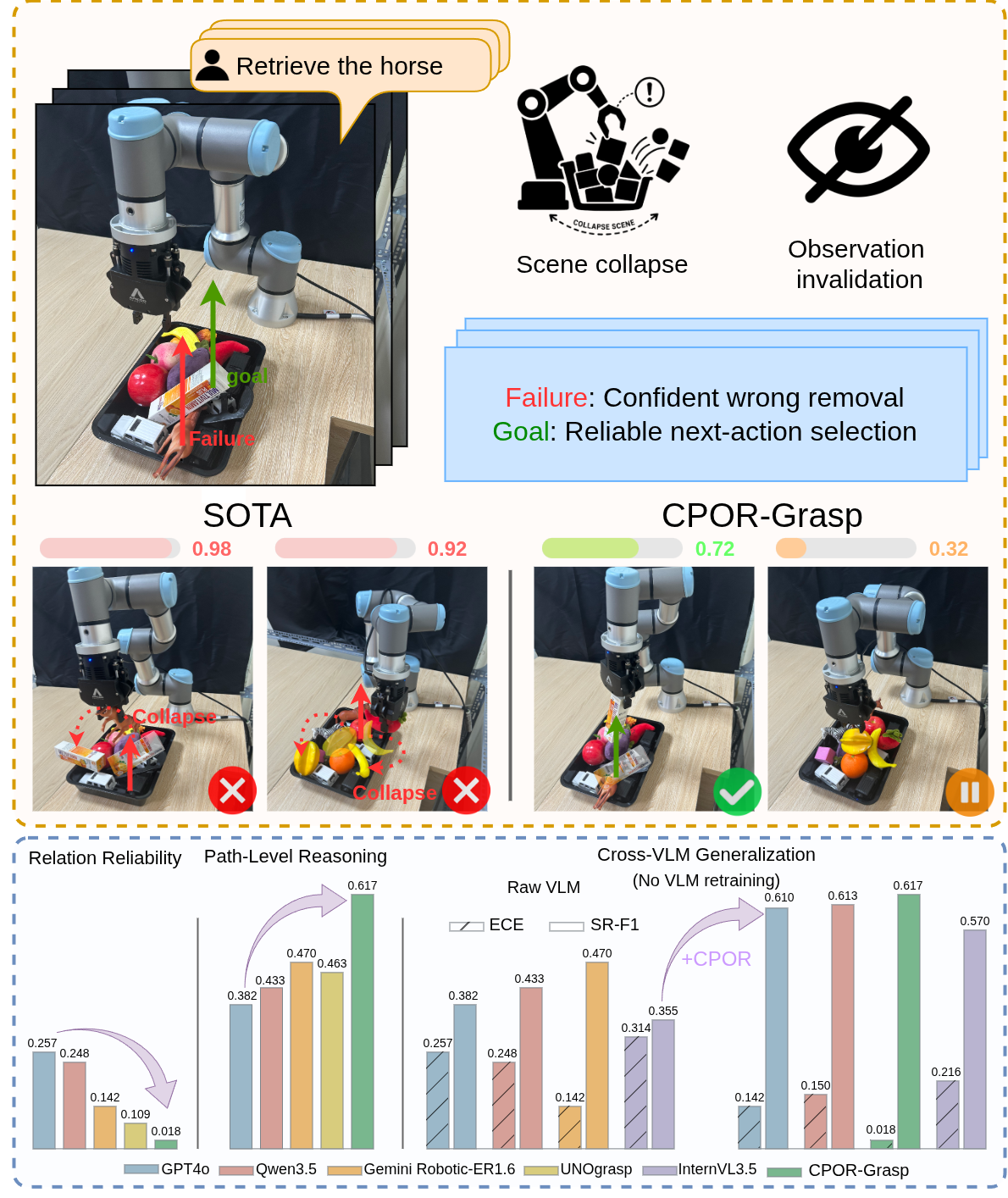}
\caption{Overconfidence versus calibrated obstruction reasoning. Top: A state-of-the-art method confidently selects an incorrect removal, causing the pile to collapse. Middle: CPOR-Grasp calibrates obstruction relations using RGB-D evidence, chooses the correct removal, and defers under high uncertainty. Bottom: A unified frozen pipeline for relation calibration, probabilistic obstruction reasoning, and transfer across VLM backbones.}
\label{fig:usecase}
\end{figure}

Retrieving a requested object from clutter is a fundamental robotic capability in warehouses, homes, and service environments~\cite{danielczuk2019mechanical}. Because targets are often blocked, the robot must decide what to remove and in which order before grasping. As illustrated in Fig.~\ref{fig:usecase}, overconfident decisions can be costly: a state-of-the-art method selects an incorrect removal with high confidence, destabilizing the scene and invalidating subsequent reasoning. In contrast, a calibrated system performs the correct removal or defers when the available evidence is insufficient.

Prior work predicts manipulation-relationship graphs~\cite{zhang2022regrad}, or asks a VLM for dependency graphs, obstruction paths, the next removal, or a grounded grasp~\cite{rabino2025modern,jiao2026obstruction}. All are judged on discrete accuracy, yet retrieval needs scores that compose as probabilities, a \emph{score-to-decision gap} with three parts. (i) Calibration: raw VLM scores are miscalibrated~\cite{xuan2025seeing}, and post-hoc calibrators~\cite{platt1999probabilistic,guo2017calibration} correct single scores, not composed graph weights. (ii) Grounding: a semantic score does not say whether RGB-D observations support the relation~\cite{yang2026fire}. (iii) Consistency: individually plausible relations can be jointly cyclic, so systems commit to one feasible graph or keep an M-best set~\cite{batra2012efficient} with no measure of what it omits.

We address these gaps with CPOR-Grasp (Calibrated Probabilistic Obstruction Reasoning for Grasping), which rescales VLM confidence with a per-scene statistic (Sec.~\ref{sec:fusion}). For grounding, reliability-gated fusion adds depth cues and amodal masks, discounting unreliable sources (Secs.~\ref{sec:local} and~\ref{sec:fusion}). For consistency, it forms a posterior over graphs that admit a removal order (Sec.~\ref{sec:global}). Marginalizing this posterior exactly is combinatorial, so CPOR-Grasp keeps the Top-$K$ graphs, bounds the omitted probability per scene, and uses this bound to certify decisions, stop adaptively, and defer (Theorem~\ref{thm:cert}, Corollary~\ref{cor:stability}, Algorithm~\ref{alg:main}). Off-the-shelf segmentation, amodal completion, and grasp execution lie outside the certified model. The main contributions can be summarized as follows:
\begin{itemize}

\item \textbf{Calibrated obstruction estimation.} We propose a scene-conditioned calibration framework for VLM confidence, combined with reliability-gated fusion of RGB-D evidence. This produces a calibrated probability for every directed obstruction relation, a capability not provided by prior obstruction-reasoning methods.

\item \textbf{Probabilistic obstruction reasoning.} 
% We formulate a posterior distribution over obstruction graphs that admit a valid removal order. By marginalizing over this distribution, the method estimates target graspability and blocker removability without committing to a single graph hypothesis.
We formulate a posterior over obstruction graphs that admit a removal order. Marginalizing over it estimates target accessibility and blocker removability without committing to a single graph hypothesis.

\item \textbf{Certified approximate inference.} 
We derive a computable per-scene total-variation bound for Top-$K$ truncation, with a decision-stability test. This test certifies when the truncated decision matches exact inference under the model, enabling adaptive stopping and deferral.
\end{itemize}

\section{Related Work}
\label{sec:related}
\paragraph{Obstruction reasoning and target retrieval in clutter}
% Manipulation-relationship networks represent access constraints as directed graphs~\cite{zhang2018visual}, and REGRAD provides RGB-D data~\cite{zhang2022regrad}. VLM-driven systems select targets or ground grasps in clutter~\cite{qian2024thinkgrasp,zhang2026gatgrasp}. D3G represents target access through a dependency graph~\cite{rabino2025modern}. UNOGrasp represents target-centered obstruction paths and defines UNOBench~\cite{jiao2026obstruction}. All are evaluated on discrete relations, paths, or actions, not on composed probabilities. Inverse-graphics reconstruction recovers geometry but no obstruction order~\cite{arriaga2026inverse}. Mechanical search and interactive perception maintain a belief over the hidden target's location or occupancy and choose removal actions~\cite{danielczuk2019mechanical}. CPOR-Grasp instead places a posterior over the directed obstruction structure and certifies that truncated decisions match exact inference under the model. Spatially supervised VLMs improve embodied spatial reasoning~\cite{chen2024spatialvlm}, whereas CPOR-Grasp calibrates and grounds a frozen VLM's scores. Amodal segmentation recovers hidden extent~\cite{back2022unseen}, occlusion ordering predicts which object lies in front~\cite{lee2022instaorder}, and FIRE-Grasp checks predictions against multimodal evidence after failure~\cite{yang2026fire}. CPOR-Grasp uses depth and amodal masks as evidence for the same directed event that the VLM scores, and discounts an unreliable source.
Manipulation-relationship networks~\cite{zhang2022regrad} represent access constraints as directed graphs. VLM-driven systems select targets or ground grasps in clutter~\cite{zhang2026gatgrasp}, D3G predicts a dependency graph~\cite{rabino2025modern}, and UNOGrasp predicts target-centered obstruction paths on UNOBench~\cite{jiao2026obstruction}. Inverse-graphics reconstruction recovers geometry but no obstruction order~\cite{arriaga2026inverse}, and mechanical search plans removals from a belief over the hidden target's location~\cite{danielczuk2019mechanical}. CPOR-Grasp instead places a posterior over the directed obstruction structure and certifies its truncated decisions. Spatially supervised VLMs improve spatial reasoning~\cite{chen2024spatialvlm}, whereas CPOR-Grasp calibrates and grounds a frozen VLM's scores. Amodal segmentation and occlusion ordering recover hidden extent and front-back order~\cite{back2022unseen,lee2022instaorder}, and FIRE-Grasp checks predictions against evidence after failure~\cite{yang2026fire}. CPOR-Grasp instead uses depth and amodal masks as evidence for the same directed event the VLM scores, discounting unreliable sources.

\paragraph{Calibration, deferral, and inference over graphs}
Verbalized VLM confidence is poorly calibrated~\cite{xuan2025seeing}. Platt and temperature scaling apply one global map~\cite{platt1999probabilistic,guo2017calibration}, input-dependent calibrators vary it with the input~\cite{tomani2022parameterized}, and CPOR-Grasp conditions its Platt-type map on a scene statistic. Selective prediction abstains at low confidence~\cite{geifman2017selective}, and language-model planners calibrate over flat plan sets and ask for help under ambiguity~\cite{liang2024introspective}. CPOR-Grasp instead defers from graph-event probabilities over order-feasible graphs with a bounded truncation error. The Top-$K$ solver follows M-best enumeration with exclusion cuts~\cite{batra2012efficient}, and CPOR-Grasp adds a bound-driven stopping rule and a decision-stability certificate. Bayesian-network structure discovery computes posteriors over directed acyclic graphs but targets edge features under decomposable scores~\cite{kuipers2017partition}. CPOR-Grasp instead needs target-accessibility and path-defined blocker-removability events, together with a per-scene certificate that avoids the normalizing constant. Prior approaches calibrate individual scores, treat geometric checks separately from semantic predictions, or enforce consistency without decision-level guarantees. CPOR-Grasp closes this score-to-decision gap.

\section{Problem Formulation}
\label{sec:formulation}
At each perception cycle, the robot observes an RGB-D image $I=(I_{rgb},I_d)$ and receives an instruction that identifies the desired object $X$. It then chooses one of three actions: grasp the target ($\mathcal{G_X}$), remove an object $o$ ($\mathcal{R_o}$), or defer ($\mathcal{D}$). Grasping the target completes retrieval, whereas removing an object changes the scene. Let $\mathcal{V}=\{X,o_1,\ldots,o_{n-1}\}$ denote the detected objects and $\cP_{\mathrm{cand}}\subseteq\{(i,j):i,j\in\mathcal{V},\ i\neq j\}$ the set of candidate directed obstruction relations. The latent obstruction state is represented by binary variables $\bfe=(e_{ij})_{(i,j)\in\cP_{\mathrm{cand}}}\in\{0,1\}^{|\cP_{\mathrm{cand}}|}$, where $e_{ij}=1$ indicates that object $j$ directly obstructs object $i$. Thus $j$ must be cleared before $i$, and indirect precedence follows by reachability along these edges. Each candidate pair carries a fused probability $p_{ij}$ for $e_{ij}=1$ (Sec.~\ref{sec:fusion}). Pairs outside $\cP_{\mathrm{cand}}$ are fixed to $e_{ij}=0$ and $p_{ij}=0$, which extends the model to all ordered pairs. All posterior inference is conditioned on the current observation context $\Omega$, comprising the RGB-D frame $I$, target instruction, detected objects $\mathcal{V}$, candidate relations $\cP_{\mathrm{cand}}$, and fused relation probabilities $\{p_{ij}\}$ from Sec.~\ref{sec:fusion}. Each decision corresponds to a single perception-to-action cycle, after which a new observation is acquired. Let $G(\bfe)$ be the directed graph on $\mathcal{V}$ with the edges selected by $\bfe$. We model one-object-at-a-time retrieval on the acyclic support $\cD=\{\bfe:G(\bfe)\text{ is acyclic}\}$. $\Reach(X,\bfe)$ contains $X$ and every vertex reachable from it by a directed path, and the out-degree $\deg_o^+(\bfe)$ counts the edges leaving $o$. For $\bfe\in\cD$, the blockers removable next are the reachable vertices with no outgoing edge:
\begin{equation}
\Free(X,\bfe)=\{o\in\Reach(X,\bfe)\setminus\{X\}:\deg_o^+(\bfe)=0\}.
\label{eq:free}
\end{equation}

With the indicator $\indic[\cdot]$, define the target event $\chi_X(\bfe)=\indic[\deg_X^+(\bfe)=0]$ and the blocker events $\chi_o(\bfe)=\indic[o\in\Free(X,\bfe)]$. For any posterior $P(\cdot\mid\Omega)$ supported on $\cD$, the decision-relevant graph-event marginals are:
\begin{equation}
q_X=\sum_{\bfe\in\cD}\chi_X(\bfe)P(\bfe\mid\Omega),
q_o=\sum_{\bfe\in\cD}\chi_o(\bfe)P(\bfe\mid\Omega).
\label{eq:marginals}
\end{equation}

% The $q_o$ do not need to sum to one, because terminal blockers on separate branches can occur together.

\noindent\textbf{Assumption 1.} \textit{Before structural conditioning, the joint edge model is the independent Bernoulli product distribution matching the fused edge probabilities.}
% \noindent\textbf{Assumption 2.} \textit{Each decision is confined to one perception-to-action cycle, so $\Omega$ describes the scene when the selected action is issued.}

% Assumption~1 defines a pre-structural reference model, not physical independence. Among joint distributions with these Bernoulli marginals, the independent product has maximum entropy. Assumption~2 limits inference to one perception-to-action cycle. The next decision uses a new observation. The quantities in \eqref{eq:marginals} are graph-implied events, not grasp-success or safety probabilities. All certificates below are conditional on $V_t$, $\cP_{\mathrm{cand}}$, and the fused probabilities $\{p_{ij}\}$.

\section{CPOR-Grasp Method}
\label{sec:method}
\begin{figure*}[!t]
\centering
\includegraphics[width=1\textwidth]{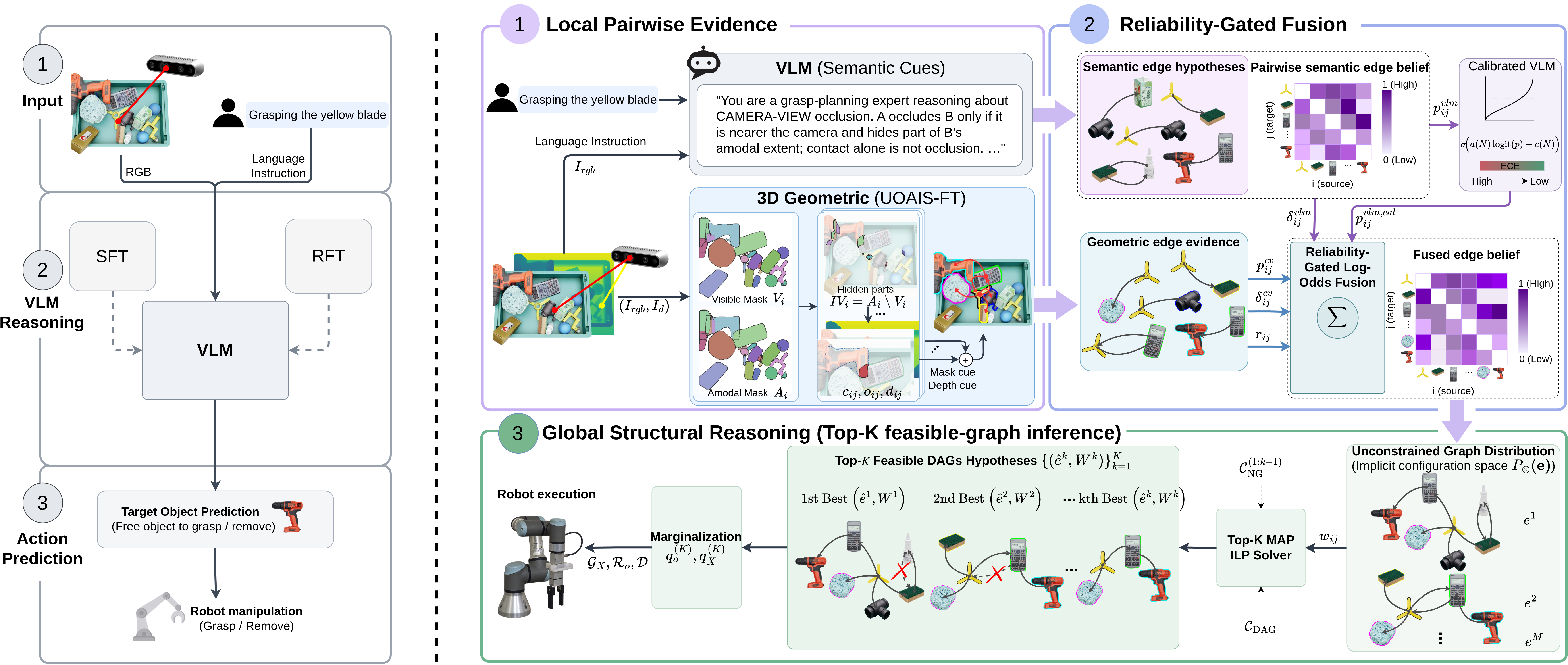}
\caption{Left: SOTA methods, right: CPOR-Grasp overview. (1) A VLM scores each ordered object pair, while the amodal segmenter and depth provide local geometric evidence. (2) Scene-conditioned calibration and reliability-gated fusion yield one probability per candidate directed relation. (3) The Top-$K$ solver enumerates the highest-weight graphs that admit a removal order, and marginalization with the truncation bound yields the grasp, remove, or defer decision.}
\label{fig:overview}
\end{figure*}

CPOR-Grasp converts an RGB-D observation into an action decision and a certified truncation bound (Fig.~\ref{fig:overview}). It segments objects, scores candidate obstruction relations with a VLM, then calibrates and fuses these scores with depth and amodal-mask evidence to obtain a probability for each directed relation. These probabilities define a distribution over order-feasible obstruction graphs, from which the $K$ most probable graphs are enumerated. Marginalization over these graphs yields the probability that the target is accessible and that each object is removable next, supporting grasp, removal, or deferral.

% CPOR-Grasp converts a single RGB-D observation into an action decision and a certified truncation bound (Fig.~\ref{fig:overview}). It segments objects, scores candidate obstruction relations with a VLM, and calibrates then fuses these scores with depth and amodal-mask evidence to obtain a probability for each directed relation. The resulting relations define a distribution over order-feasible obstruction graphs, from which the $K$ most probable graphs are enumerated. Marginalization over these graphs yields the probability that the target is graspable and that each object is the next removable blocker, enabling grasp, removal, or deferral decisions.

\subsection{Local Pairwise Evidence}
\label{sec:local}
Two sources score the same directed event $e_{ij}=1$ on separate candidate sets. Let $\cP_{vlm}$ be the set of ordered pairs the VLM scored, with raw score $p^{vlm}_{ij}$ computed from normalized yes/no logits. Let $\cP_{cv}$ be the pairs admitted by a fixed geometric rule. A pair enters $\cP_{cv}$ when the visible mask of $j$ overlaps the hidden region of object $i$ or a clearance ring of $\rho$ pixels around it, and the overlap carries valid depth on at least half of its pixels. Fusion operates on $\cP_{\mathrm{cand}}=\cP_{vlm}\cup\cP_{cv}$, with indicators $\delta^{vlm}_{ij},\delta^{cv}_{ij}\in\{0,1\}$ recording membership. A pair the VLM did not score has $\delta^{vlm}_{ij}=0$ and carries no semantic evidence, whereas a pair scored low carries negative evidence.

\textit{(a) Mask cues.}
The geometric branch uses UOAIS-FT, our domain-adapted amodal segmenter obtained by fine-tuning UOAIS~\cite{back2022unseen} on 2,000 synthetic images sampled separately from the 1,800-image benchmark test set. The VLM answers in badge numbers, and each badge point is matched to a mask, first inside a visible mask, then inside an amodal mask, then to the nearest centroid within 120 pixels. An amodal mask estimates the full extent of an object, hidden part included. Let $V_i$ and $A_i$ be the post-processed visible and amodal masks of object $i$. Hidden overlap $o_{ij}$ measures how much of $i$ lies hidden under $j$. Local clearance $c_{ij}$ is the fraction of a ring around $i$ that $j$ occupies:
\begin{equation}
o_{ij}=\frac{|IV_i\cap V_j|}{\max(|A_i|,1)},\quad
c_{ij}=\frac{|R_i^{\mathrm{clr}}\cap V_j|}{\max(|R_i^{\mathrm{clr}}|,1)},
\label{eq:mask-support}
\end{equation}
where $IV_i=A_i\setminus V_i$ is the hidden part of $i$ and $R_i^{\mathrm{clr}}=\operatorname{Dilate}(A_i,\rho)\setminus A_i$ is a ring of width $\rho$ around $A_i$.

\textit{(b) Depth cue.} Mask cues do not say which object lies in front. Depth $d_{ij}=r_{ij}\tanh\!\left(\frac{z_i-z_j}{\sigma_z}\right)$ supplies that cue, and a linear score combines all three:
\begin{equation}
\begin{gathered}
u^{cv}_{ij}=b^{cv}+\kappa_o(o_{ij}-o_\star)+\kappa_c(c_{ij}-c_\star)+\kappa_zd_{ij},
\end{gathered}
\label{eq:geometry-score}
\end{equation}
where $z_i$ is the median valid depth over $V_i$, and smaller depth is closer to the camera. The scale $\sigma_z>0$ sets the separation at which the ordering cue saturates. The factor $r_{ij}=\min(r_i^z,r_j^z)\in[0,1]$ is the smaller valid-depth fraction, where $r_i^z$ is the fraction of $V_i$ with valid depth. If either median depth is unavailable, $d_{ij}=0$. The coefficients $\kappa_o,\kappa_c,\kappa_z$ are fixed and nonnegative. The remaining parameters $b^{cv},o_\star,c_\star$ and the radius $\rho$ are fixed before test evaluation. Let $\sigma(u)=1/(1+e^{-u})$ be the logistic sigmoid, distinct from the depth scale $\sigma_z$. It maps the score to a bounded confidence $p^{cv}_{ij}=\sigma(u^{cv}_{ij})$, not a calibrated posterior.

\subsection{Reliability-Gated Fusion}
\label{sec:fusion}
\textit{(a) Scene-conditioned calibration.} Raw VLM scores are miscalibrated, and the miscalibration varies with the scene, so we rescale them with a scene statistic. Let $N_s^{vlm}=|\cP_{vlm,s}|$ count the VLM-scored pairs in scene $s$. Its standardized log count is $\phi_s=[\log(1+N_s^{vlm})-\mu_\phi]/\sigma_\phi$, where $\mu_\phi$ and $\sigma_\phi>0$ are frozen statistics of the calibration split. With $0<\varepsilon_{vlm}<1/2$, the semantic score is calibrated by
\begin{equation}
% \begin{aligned}
\scriptsize
p^{\mathrm{clip}}_{ij}=\clip(p^{vlm}_{ij},\varepsilon_{vlm},1-\varepsilon_{vlm}),
p^{vlm,cal}_{ij}=\sigma\!\left(a_s\logit p^{\mathrm{clip}}_{ij}+c_s\right),
% \end{aligned}
\label{eq:calibration}
\end{equation}
where $a_s=\exp[\clip(\alpha_0+\alpha_N\phi_s,-5,5)]$ and $c_s=c_0+c_N\phi_s$ are scene-dependent Platt scale and bias, and $(\alpha_N,c_N)$ controls their dependence on $\phi_s$. Since $a_s>0$, the map changes score magnitude without reversing within-scene order. With $\Theta_{\mathrm{cal}}=(\alpha_0,\alpha_N,c_0,c_N)$ and $\ell_n^{\mathrm{cal}}$ for the Bernoulli log-likelihood of the $n$th of $N_{\mathrm{cal}}$ labelled calibration pairs, we fit
\begin{equation}
\Theta_{\mathrm{cal}}^\star=\arg\min_{\Theta_{\mathrm{cal}}}
-\frac{1}{N_{\mathrm{cal}}}\sum_n\ell_n^{\mathrm{cal}}
+\frac{\lambda_{\mathrm{cal}}}{2}(\alpha_N^2+c_N^2),
\label{eq:cal-loss}
\end{equation}
where $\lambda_{\mathrm{cal}}\ge0$, the penalty regularizes only
the scene-conditioning coefficients $\alpha_N$ and $c_N$.

\textit{(b) Gated fusion.}
Let $\pi_0\in(0,1)$ be the positive-edge rate on the fitting split. The centered log-odds are $h^m_{ij}=\logit p^m_{ij}-\logit\pi_0$ for $m\in\{vlm,cv\}$, where $\logit p=\log[p/(1-p)]$. The VLM term uses $p^{vlm,cal}_{ij}$ and the geometric term $p^{cv}_{ij}$. %For an unavailable source, set $h^m_{ij}=0$, and set $r_{ij}=0$ when $\delta^{cv}_{ij}=0$. 
In the Bayes reference case of Appendix\ref{app:reference}, centered log-odds add exactly. CPOR-Grasp instead learns the source contributions:
\begin{equation}
g_{ij}=\gamma+\delta^{vlm}_{ij}\beta_{vlm}h^{vlm}_{ij}
+\delta^{cv}_{ij}\beta_{cv}r_{ij}h^{cv}_{ij}.
\label{eq:fusion}
\end{equation}

We call \eqref{eq:fusion} reliability-gated fusion, because $r_{ij}$ gates the geometric source by its depth support. With $\widetilde p_{ij}=\sigma(g_{ij})$, the fusion parameters $\Theta_{\mathrm{fus}}=(\gamma,\beta_{vlm},\beta_{cv})$ are fit by
\begin{equation}
\Theta_{\mathrm{fus}}^\star=
\argmin_{\substack{\gamma\in\mathbb R,\\\beta_{vlm},\beta_{cv}\ge0}}
-\frac{1}{N_{\mathrm{fus}}}\sum_n\ell_n^{\mathrm{fus}}
+\frac{\lambda_{\mathrm{fus}}}{2}(\beta_{vlm}^2+\beta_{cv}^2),
\label{eq:fusion-loss}
\end{equation}
where $\ell_n^{\mathrm{fus}}$ is the Bernoulli log-likelihood of the unclipped $\widetilde p_n$ over the $N_{\mathrm{fus}}$ labelled fusion pairs. With $\lambda_{\mathrm{fus}}\ge0$, the intercept is unregularized. The sign constraints let an unhelpful source fall to zero weight without reversing its evidence. Structural inference uses $p_{ij}=\clip(\widetilde p_{ij},\zeta,1-\zeta)$ with $0<\zeta<1/2$.

\subsection{Global Structural Reasoning}
\label{sec:global}
\textit{(a) Feasible-graph posterior.}
The fused $p_{ij}$ describes each relation alone, so plausible relations may together form a cycle that no removal order realizes. We therefore restrict the joint edge model to $\cD$. Under Assumption~1, the pre-structural product model is:
\begin{equation}
P_\otimes(\bfe\mid\Omega)=\prod_{(i,j)\in\cP_{\mathrm{cand}}}
p_{ij}^{e_{ij}}(1-p_{ij})^{1-e_{ij}},
\label{eq:weight}
\end{equation}
and conditioning it on acyclicity gives the feasible-directed acyclic graph (DAG) posterior:
\begin{equation}
\PD(\bfe\mid\Omega)=\frac{\indic[\bfe\in\cD]P_\otimes(\bfe\mid\Omega)}{Z},
Z=\sum_{\bfe\in\cD}P_\otimes(\bfe\mid\Omega).
\label{eq:graph-posterior}
\end{equation}
The rejected mass $\mu=1-Z$ measures conflict between the product model and acyclicity. %, not physical failure. 
Every candidate-pair $p_{ij}$ lies in $[\zeta,1-\zeta]$. Hence $\mu>0$ exactly when the candidate digraph $(\mathcal{V},\cP_{\mathrm{cand}})$, the graph of all candidate edges, contains a directed cycle, and $\mu=0$ otherwise. The fused $p_{ij}$ parameterizes the pre-structural model and need not equal $\PD(e_{ij}{=}1\mid\Omega)$.

\textit{(b) Highest-weight Top-$K$.}
Exact evaluation of \eqref{eq:marginals} under \eqref{eq:graph-posterior} is combinatorial. Let $S_0=\emptyset$, and write $w_{ij}=\logit p_{ij}$ and $\ell_0=\sum_{(i,j)\in\cP_{\mathrm{cand}}}\log(1-p_{ij})$. Then $\log P_\otimes(\bfe\mid\Omega)=\ell_0+\sum w_{ij}e_{ij}$, and the $k^{th}$ retained graph is
\begin{equation}
\widehat\bfe^{(k)}\in\arg\max_{\bfe\in\cD\setminus S_{k-1}}
\sum_{(i,j)\in\cP_{\mathrm{cand}}}w_{ij}e_{ij},
\label{eq:topk}
\end{equation}
where $S_k=S_{k-1}\cup\{\widehat\bfe^{(k)}\}$. Appendix\ref{app:ilp} gives the cuts that implement \eqref{eq:topk}. Let $Z_K=\sum_{\bfe\in S_K}P_\otimes(\bfe\mid\Omega)$ be the retained mass and $W^{(k)}:=P_\otimes\!\left(\widehat\bfe^{(k)}\mid\Omega\right)/Z_K$ the normalized weight of the $k^{th}$ graph. The retained posterior is:
\begin{equation}
\small
\PK(\bfe\mid\Omega)=\frac{\indic[\bfe\in S_K]P_\otimes(\bfe\mid\Omega)}{Z_K}
=\PD(\bfe\mid\Omega,\bfe\in S_K).
\label{eq:pk}
\end{equation}

Replacing $P$ in \eqref{eq:marginals} with \eqref{eq:pk} gives $\widehat q_X^{(K)}=\sum_{k\le K}W^{(k)}\chi_X(\widehat\bfe^{(k)})$ and likewise $\widehat q_o^{(K)}$. Thus $K=1$ is the MAP plug-in, and $K>1$ retains several feasible structures. Unhatted $q_X$ and $q_o$ denote the exact marginals under $\PD$, and hatted ones their Top-$K$ estimates.

\textit{(c) Computable truncation certificate.}
The omitted mass can be bounded without evaluating $Z$. Let $\cR_2=\{\bfe:e_{ij}+e_{ji}\le1\text{ for every unordered pair }\{i,j\}\}$ collect the configurations with no two-cycle, a pair of opposite edges. Every acyclic graph lies in $\cR_2$, so
\begin{equation}
Z\le\overline Z:=P_\otimes(\cR_2\mid\Omega)=\prod_{i<j}(1-p_{ij}p_{ji}),
\label{eq:zbar}
\end{equation}
under the full-pair convention of Sec.~\ref{sec:formulation}. Longer directed cycles can make \eqref{eq:zbar} conservative, but cannot invalidate it. The certificate uses total variation, the largest difference in probability that two distributions assign to any event.

\theoremhead{thm:cert}{Top-$K$ truncation certificate}
\textit{Let $1\le K\le|\cD|$ and let $S_K\subseteq\cD$ contain $K$ distinct feasible graphs. With $\dTV(P,Q)=\tfrac12\sum_{\bfe\in\cD}|P(\bfe)-Q(\bfe)|$,}
\begin{equation}
\dTV(\PD,\PK)=1-\frac{Z_K}{Z}
\le\epsilon_K:=1-\frac{Z_K}{\overline Z}.
\label{eq:certificate}
\end{equation}

Consequently, for every $f:\cD\to[0,1]$, $|\mathbb E_{\PD}[f]-\mathbb E_{\PK}[f]|\le\epsilon_K$. In particular, the same bound holds for $q_X$ and every $q_o$. Because $\overline Z$ is fixed for a scene, the highest-weight construction, which maximizes $Z_K$, also minimizes $\epsilon_K$. Conditioning removes the mass $\mu$, whereas $\epsilon_K$ bounds only the truncation error $1-Z_K/Z$. If enumeration is proven exhausted, then $Z_K=Z$ and the true truncation error is zero. 
% Algorithm~\ref{alg:main} therefore sets $\mathrm{exact}\gets\mathrm{true}$; the conservative analytic bound $\epsilon_K$ in \eqref{eq:certificate} need not vanish.

\textit{(d) Decision and adaptive stopping.}
The same $\epsilon_K$ yields sufficient conditions under which the approximate scores are guaranteed to reproduce the exact decision. Let $\cA_{\mathrm{exec}}=\{\mathcal{G_X}\}\cup\{\mathcal{R_o}:o\in\mathcal{V}\setminus\{X\}\}$ be the executable actions and $\cA=\cA_{\mathrm{exec}}\cup\{\mathcal{D}\}$ the decision set. Set $s_{\mathcal{G_X}}=\widehat q_X^{(K)}$ and $s_{\mathcal{R_o}}=\widehat q_o^{(K)}$. With $s_{(1)}$ the largest approximate executable-action score and a fixed threshold $\tau\in[0,1]$, the approximate policy is
\begin{equation}
\small
\widehat a=\begin{cases}
\arg\max_{a\in\cA_{\mathrm{exec}}}s_a,&s_{(1)}>\tau,\\
\mathcal{D},&s_{(1)}\le\tau,
\end{cases} 
\widehat F=\{o:\widehat q_o^{(K)}>\tau\}.
\label{eq:policy}
\end{equation}

Define the exact scores $s_{\mathcal{G_X}}^\star=q_X$ and $s_{\mathcal{R_o}}^\star=q_o$. Let $a^\star\in\cA$ follow the same threshold rule and tie-breaking on the exact scores, and let $F^\star=\{o:q_o>\tau\}$. 

\corollaryhead{cor:stability}{Operational stability}
\textit{Let $s_{(2)}$ be the second-largest approximate executable-action
score, with $s_{(2)}:=-\infty$ when $|\cA_{\mathrm{exec}}|=1$, and define}
$\delta_{\mathrm{rank}}=s_{(1)}-s_{(2)},
% \delta_{\mathrm{act}}=|s_{(1)}-\tau|,
\delta_o=|\widehat q_o^{(K)}-\tau|.$
% \label{eq:stability}
% \end{equation}
\textit{
(a) If $s_{(1)}-\epsilon_K>\tau$ and either
$|\cA_{\mathrm{exec}}|=1$ or
$\delta_{\mathrm{rank}}>2\epsilon_K$,
then $\widehat a=a^\star$.
% If $\delta_{\mathrm{act}}>\epsilon_K$ implies
% $(\widehat a=\mathcal D)\Leftrightarrow(a^\star=\mathcal D)$.
(b) If $s_{(1)}+\epsilon_K\le\tau$, then
$\widehat a=a^\star=\mathcal D$.
(c) For each object $o$,
$\delta_o>\epsilon_K$ implies
$o\in\widehat F\Leftrightarrow o\in F^\star$.
}
% \textit{Then $\delta_{\mathrm{rank}}>2\epsilon_K$ implies $\widehat a=a^\star$ whenever $\widehat a,a^\star\in\cA_{\mathrm{exec}}$, , and $\delta_o>\epsilon_K$ implies $o\in\widehat F\Leftrightarrow o\in F^\star$.}

% All three tests use the same $\epsilon_K$. The one-sided condition $s_{(1)}+\epsilon_K\le\tau$ certifies $\widehat a=a^\star=\mathcal{D}$, and $\widehat F=F^\star$ when the blocker-margin test holds for every candidate object. These guarantees concern agreement with exact inference under \eqref{eq:graph-posterior}, not perception, model correctness, grasp success, execution safety, or task-level utility optimality.
% \begin{remark}[Scope of the guarantees]
% All guarantees are conditional on the retained objects $\mathcal{V}$, the candidate set $\cP_{\mathrm{cand}}$, and the fused probabilities $\{p_{ij}\}$. They bound the discrepancy between Top-$K$ and exact inference under \eqref{eq:graph-posterior}. They do not cover relations missed by candidate generation. The marginals $q_X$ and $q_o$ are graph-event probabilities, not grasp-success or safety probabilities. Calibrated $p_{ij}$ therefore do not imply calibrated $q_X$ or $q_o$. Assumption~1 is a maximum-entropy reference model, not a claim of physical independence.
% \end{remark}

Algorithm~\ref{alg:main} stops early when test (a) or (b) holds. It certifies the blocker set \(\widehat F\) when exact inference is reached or when test (c) holds for every object. Certification costs one $O(n^2)$ evaluation of $\overline Z$ and an $O(1)$ update of $Z_K$ and $\epsilon_K$ per retained graph. A certified exit satisfies test (a) or (b) of Corollary~\ref{cor:stability}, whereas a tolerance or $K_{\max}$ exit certifies nothing. If exact$=\mathrm{true}$, the returned marginals are exact. 
% \textcolor{blue}{a tolerance or \(K_{\max}\) exit provides no decision certificate, although \(\widehat F\) may still be certified by \(\mathrm{cert}_F\).}

\begin{algorithm}[t]
\caption{CPOR-Grasp Algorithm}
\label{alg:main}
\scriptsize
\algrenewcommand\algorithmicrequire{\textbf{Input:}}
\algrenewcommand\algorithmicensure{\textbf{Output:}}

\begin{algorithmic}[1]

% \Require RGB-D frame $I$, instruction $p$, target $X$, objects $\mathcal{V}$ \\
% frozen perception, calibration, and fusion parameters;
% $K_{\max}$, $\epsilon_{\mathrm{tol}}$, $\tau$
% \Ensure $\widehat a$, $\widehat F$, $\mathrm{cert}_F$, $\widehat{\mathbf q}^{(K)}$,
% $K$, $\epsilon_K$

% \Require RGB-D frame $I$, instruction $p$, target $X$, objects $\mathcal V$;\\
% frozen UOAIS-FT and VLM models;
% $\Theta_{\rm cal}^\star$, $\Theta_{\rm fus}^\star$;\\
% $\rho,\sigma_z,\kappa_o,\kappa_c,\kappa_z,b^{cv},o_\star,c_\star,
% \mu_\phi,\sigma_\phi,\varepsilon_{vlm},\pi_0,\zeta$;\\
% $K_{\max}\ge1$, $\epsilon_{\rm tol}\in[0,1]$, $\tau\in[0,1]$$p$

\Require frame $I$, instruction $p$, target $X$, objects $\mathcal V$, 
frozen UOAIS-FT and VLM models; \\
fitted $\Theta_{\rm cal}^\star$, $\Theta_{\rm fus}^\star$; 
$\rho,\sigma_z,\kappa_o,\kappa_c,\kappa_z,b^{cv},o_\star,c_\star,
\mu_\phi,\sigma_\phi,\varepsilon_{vlm},\pi_0,\zeta$;\\
$K_{\max}\ge1$, $\epsilon_{\rm tol}\in[0,1]$, $\tau\in[0,1]$
\Ensure $\widehat a$, $\widehat F$, $\mathrm{cert}_F$,
$\widehat{\mathbf q}^{(K)}$, $K$, $\epsilon_K$, exact

\State $\{V_i,A_i\}\gets\textsc{UoaisFt}(I)$,
\quad
$(\cP_{vlm},\{p^{vlm}_{ij}\})
\gets\textsc{Vlm}(I_{rgb},p,\mathcal{V})$

\State $\phi_s\gets
[\log(1+|\cP_{vlm}|)-\mu_\phi]/\sigma_\phi$

\State $\cP_{cv},\{p^{cv}_{ij},r_{ij}\}
\gets\eqref{eq:mask-support}$ to \eqref{eq:geometry-score}

\State $\{p^{vlm,cal}_{ij}\}_{(i,j)\in\cP_{vlm}}
\gets\eqref{eq:calibration}$

\State $\cP_{\mathrm{cand}}\gets
\cP_{vlm}\cup\cP_{cv}$

\ForAll{$(i,j)\in\cP_{\mathrm{cand}}$}
    \State $g_{ij}\gets\eqref{eq:fusion}$ using available sources
    \State $p_{ij}\gets
    \clip(\sigma(g_{ij}),\zeta,1-\zeta)$,
    \quad $w_{ij}\gets\logit p_{ij}$
\EndFor

\State $\ell_0\gets
\sum_{(i,j)\in\cP_{\mathrm{cand}}}\log(1-p_{ij})$,
\quad
$\log\overline Z\gets
\sum_{i<j}\log(1-p_{ij}p_{ji})$

\State $\cC_{\mathrm{NG}}\gets\emptyset$,
$K\gets0$,
$\log Z_K\gets-\infty$,
exact$\gets\mathrm{false}$

\While{$K<K_{\max}$}

    \State $(\bfe,v)\gets
    \textsc{NextFeasibleDag}(\cC_{\mathrm{NG}})
    $ using \eqref{eq:topk} \Comment{$v=\sum_{(i,j)}w_{ij}e_{ij}$}

    \If{solver proves infeasibility}
        \State exact$\gets\mathrm{true}$ \Comment{support exhausted; true truncation error is zero}
        \State \textbf{break}
    \EndIf

    \State $K\gets K+1$,
    $\widehat\bfe^{(K)}\gets\bfe$,
    $\ell_K\gets\ell_0+v$

    \State add $\textsc{NoGood}(\bfe)$ to $\cC_{\mathrm{NG}}$

    \State $\log Z_K\gets
    \Lse(\log Z_K,\ell_K)$ \Comment{$\Lse(a,b)=\log(e^a+e^b)$}

    \State $W^{(k)}\gets\exp(\ell_k-\log Z_K)$
    for $k=1,\ldots,K$

    \State $\widehat{\mathbf q}^{(K)}\gets
    \sum_{k=1}^{K}W^{(k)}\boldsymbol\chi(\widehat\bfe^{(k)})$
    \Comment{$\boldsymbol\chi=(\chi_X,(\chi_o)_{o\neq X})$}

    \State $\epsilon_K\gets
    1-\exp(\log Z_K-\log\overline Z)$

    \State form action scores $s_a$ from $\widehat{\mathbf q}^{(K)}$

    \If{$s_{(1)}+\epsilon_K\le\tau$}
        \State \textbf{break} \Comment{$\mathcal{D}$ certified, test (b)}
    \EndIf

    \If{$s_{(1)}-\epsilon_K>\tau$
    and $(|\cA_{\mathrm{exec}}|=1$ or $s_{(1)}-s_{(2)}>2\epsilon_K)$}
        \State \textbf{break} \Comment{action certified, test (a)}
    \EndIf

    \If{$\epsilon_K\le\epsilon_{\mathrm{tol}}$}
        \State \textbf{break} \Comment{tolerance only}
    \EndIf

\EndWhile

\State $\widehat a,\widehat F\gets\eqref{eq:policy}$

\State $\mathrm{cert}_F\gets
\mathrm{exact}\ \lor\
\big[\forall o\in\mathcal{V}\setminus\{X\}:
|\widehat q_o^{(K)}-\tau|>\epsilon_K\big]$
\Comment{exact or test (c)}

\Return $\widehat a,\widehat F,\mathrm{cert}_F,
\widehat{\mathbf q}^{(K)},K,\epsilon_K,$ exact

\end{algorithmic}
\end{algorithm}

\section{Experimental Evaluation}
\label{sec:experiments}
\subsection{Experimental Protocol}
\label{sec:exp-protocol}

% \noindent\textit{Benchmarks and baselines.}
We evaluate on a held-out subset of UNOBench~\cite{jiao2026obstruction}, comprising 1,800 synthetic and 838 real-world target-retrieval scenes. Baselines include UNOGrasp~\cite{jiao2026obstruction}, Gemini Robotics-ER-1.6, GPT-4o, Qwen3.5-9B, and InternVL3.5-14B; Gemma3-12B is additionally evaluated in the cross-VLM study. CPOR-Grasp uses Gemini Robotics-ER-1.6 as its frozen semantic backbone. Calibration and fusion are fitted on separate development splits disjoint from all test sets.
% All baselines operate on RGB images only. 
% \noindent\textit{Task metrics.}
% Following the UNOBench evaluation protocol~\cite{jiao2026obstruction}, we report relation precision, recall, and $F_1$ (SR-P/R/$F_1$), Multi-Path Normalized Edit Distance (MP-NED) for complete obstruction paths, and object precision, recall, and $F_1$ (OR-P/R/$F_1$) for next-obstructor prediction.
SR-P/R/$F_1$ are the precision, recall, and $F_1$ of predicted directed obstruction relations against the ground-truth graph. MP-NED is the multi-path normalized edit distance between predicted and ground-truth obstruction paths, and lower is better. OR-P/R/$F_1$ are the precision, recall, and $F_1$ of the predicted next-removable blockers.

% Calibration and fusion parameters are fit on a held-out synthetic split of 550 retrieval cases over 473 images, case-disjoint from the test set, with a scene-disjoint validation part. The fitted calibrators are reused unchanged for real-world evaluation and for every backbone. Fixed settings are $\zeta=10^{-9}$, $\varepsilon_{vlm}=10^{-6}$, $\rho=8$ pixels, $\sigma_z=60$\,mm, $\kappa=12$ with $o_\star=0.15$, $\lambda=1.0$ for the semantic calibrator and $0.001$ for the geometric one, $K_{\max}=256$, $\epsilon_{\mathrm{tol}}=0.05$, and 10 equal-width ECE bins. Top-$K$ enumeration uses the CBC solver through PuLP with one thread and a 30\,s limit per solve, and every eligible scene stays below the exact-enumeration cutoff of 20 candidate pairs. Semantic scores come from one prompt at temperature 0 that shows the badge-annotated image, the object list, and the geometric hypothesis table, and elicits a verbalized probability from 0 to 100 per accepted relation under an anchored rubric. Decoding shifts each fused probability by the edge prior $\tau_{edge}=0.09$, reads relations and paths from the MAP graph, and thresholds object marginals at 0.5 [TBD: $\tau_{edge}$ was selected on the test split, re-select on the fitting split or disclose]. Tables~\ref{tab:path-bench} and~\ref{tab:obj-bench} use forced decisions without deferral, that is $\tau=0$. Code and the fused score logs are available at \url{https://github.com/pairs-lab/cpor-grasp}.

\subsection{Obstruction Reasoning Results}
\label{sec:main-results}

% UNOBench defines difficulty from obstruction-graph depth and the number of reasoning paths~\cite{jiao2026obstruction}; Hard scenes contain the most complex removal structures. CPOR-Grasp path-level numbers are read from the fused probabilities before structural inference.

\begin{table*}[!t]
\centering
\caption{Path-level reasoning on UNOBench synthetic and real-world test sets. \textbf{Best}/\underline{second-best} values are bold/underlined.}
\label{tab:path-bench}
\vspace{-2.5mm}
\scriptsize
\setlength{\tabcolsep}{0pt} 
\renewcommand{\arraystretch}{1.0}

\begin{tabular*}{\textwidth}{@{\extracolsep{\fill}}l l c cccc cccc cccc@{}}
\toprule
% \multicolumn{14}{l}{\textbf{A. Synthetic test set}}\\
% \midrule
& \multirow{2}{*}{Method} & \multicolumn{1}{c}{Easy} & \multicolumn{4}{c}{Medium} & \multicolumn{4}{c}{Hard} & \multicolumn{4}{c}{Overall} \\
\cmidrule(lr){3-3}\cmidrule(lr){4-7}\cmidrule(lr){8-11}\cmidrule(lr){12-15}
& & MP-NED$\downarrow$ & $P\uparrow$ & $R\uparrow$ & $F_1\uparrow$ & MP-NED$\downarrow$ & $P\uparrow$ & $R\uparrow$ & $F_1\uparrow$ & MP-NED$\downarrow$ & $P\uparrow$ & $R\uparrow$ & $F_1\uparrow$ & MP-NED$\downarrow$ \\
\midrule
\multirow{6}{*}{\rotatebox{90}{\textbf{A. Synthetic}}} & Gemini Robotics-ER-1.6~\cite{gemini_robotics_er_16} & \underline{0.0650} & 0.5492 & 0.5633 & 0.5472 & 0.2953 & 0.5611 & \underline{0.3421} & \underline{0.3939} & \underline{0.5505} & 0.5551 & \underline{0.4527} & \underline{0.4705} & \underline{0.3036} \\
& Qwen3.5~\cite{qwen35}                              & 0.1946 & 0.5700 & 0.5608 & 0.5608 & 0.3187 & \underline{0.5783} & 0.2172 & 0.3062 & 0.6307 & \underline{0.5741} & 0.3890 & 0.4335 & 0.3813 \\
& GPT-4o~\cite{openai2024gpt4o}                      & 0.1326 & 0.5117 & 0.4942 & 0.4989 & 0.3356 & 0.5200 & 0.1843 & 0.2646 & 0.6672 & 0.5158 & 0.3392 & 0.3817 & 0.3784 \\
& UNOGrasp~\cite{jiao2026obstruction}                 & 0.0699 & \underline{0.6078} & \underline{0.6000} & \underline{0.5974} & \underline{0.2519} & 0.5224 & 0.2631 & 0.3276 & 0.5707 & 0.5651 & 0.4316 & 0.4625 & 0.4113 \\
& InternVL3.5~\cite{wang2025internvl3}            & 0.1783 & 0.4700 & 0.4550 & 0.4594 & 0.3507 & 0.4772 & 0.1779 & 0.2505 & 0.6708 & 0.4736 & 0.3164 & 0.3549 & 0.3999 \\
% \midrule
& \textbf{Our CPOR-Grasp}                                      & \textbf{0.0335} & \textbf{0.6725} & \textbf{0.7450} & \textbf{0.6856} & \textbf{0.2134} & \textbf{0.6578} & \textbf{0.5398} & \textbf{0.5491} & \textbf{0.4273} & \textbf{0.6652} & \textbf{0.6424} & \textbf{0.6174} & \textbf{0.2247} \\

\midrule
% \multicolumn{14}{l}{\textbf{B. Real-world test set}}\\
% \midrule
\multirow{6}{*}{\rotatebox{90}{\textbf{B. Real-world}}} & Gemini Robotics-ER-1.6~\cite{gemini_robotics_er_16} & \underline{0.1060} & 0.5334 & 0.5128 & 0.5134 & 0.3126 & \underline{0.5036} & \underline{0.2935} & \underline{0.3424} & \underline{0.6039} & 0.5202 & 0.4157 & 0.4377 & \underline{0.3213} \\
& Qwen3.5~\cite{qwen35}                              & 0.2603 & 0.6467 & 0.6028 & 0.6158 & 0.2731 & 0.5007 & 0.2003 & 0.2769 & 0.6520 & \underline{0.5821} & \underline{0.4247} & \underline{0.4658} & 0.3761 \\
& GPT-4o~\cite{openai2024gpt4o}                      & 0.2144 & 0.5400 & 0.4972 & 0.5109 & 0.3242 & 0.3782 & 0.1330 & 0.1911 & 0.7120 & 0.4684 & 0.3360 & 0.3694 & 0.3950 \\
& UNOGrasp~\cite{jiao2026obstruction}                 & 0.1239 & \underline{0.6772} & \underline{0.6433} & \underline{0.6506} & \underline{0.2483} & 0.4040 & 0.1734 & 0.2297 & 0.6629 & 0.5406 & 0.4083 & 0.4402 & 0.3450 \\
& InternVL3.5~\cite{wang2025internvl3}                & 0.2361 & 0.5383 & 0.4911 & 0.5056 & 0.3439 & 0.3634 & 0.1299 & 0.1866 & 0.7051 & 0.4609 & 0.3313 & 0.3644 & 0.4078 \\
% \midrule
& \textbf{Our CPOR-Grasp}                                      & \textbf{0.0744} & \textbf{0.6960} & \textbf{0.7022} & \textbf{0.6842} & \textbf{0.2082} & \textbf{0.5609} & \textbf{0.4515} & \textbf{0.4583} & \textbf{0.4971} & \textbf{0.6362} & \textbf{0.5913} & \textbf{0.5843} & \textbf{0.2423} \\
\bottomrule
\multicolumn{14}{l}{\textit{$^*$Note: Easy scenes contain no obstruction relations.}} \\
\end{tabular*}
\vspace{-1.5mm}
\end{table*}

\begin{table*}[!t]
\centering
\caption{Object-level next-obstructor prediction on UNOBench dataset. \textbf{Best}/\underline{second-best} values are bold/underlined.}
\label{tab:obj-bench}
\vspace{-2.5mm}
\scriptsize 
\setlength{\tabcolsep}{-1pt} 
\renewcommand{\arraystretch}{1.0}

\begin{tabular*}{\textwidth}{@{\extracolsep{\fill}}l l l cc ccc ccc ccc@{}}
\toprule
% \multicolumn{13}{l}{\textbf{Object-level next-obstructor prediction results}}\\
% \midrule
& \multirow{2}{*}{Method} & \multicolumn{3}{c}{Easy} & \multicolumn{3}{c}{Medium} & \multicolumn{3}{c}{Hard} & \multicolumn{3}{c}{Overall} \\
\cmidrule(lr){3-5}\cmidrule(lr){6-8}\cmidrule(lr){9-11}\cmidrule(lr){12-14}
& & $P\uparrow$ & $R\uparrow$ & $F_1\uparrow$ & $P\uparrow$ & $R\uparrow$ & $F_1\uparrow$ & $P\uparrow$ & $R\uparrow$ & $F_1\uparrow$ & $P\uparrow$ & $R\uparrow$ & $F_1\uparrow$ \\
\midrule
\multirow{6}{*}{\rotatebox{90}{\textbf{A. Synthetic}}} & Gemini Robotics-ER-1.6~\cite{gemini_robotics_er_16} & \underline{0.9233} & \underline{0.9283} & \underline{0.9246} & 0.6525 & 0.6278 & 0.6302 & 0.5199 & 0.4686 & 0.4755 & 0.6986 & 0.6749 & 0.6767 \\
& Qwen3.5~\cite{qwen35}                              & 0.6792 & 0.6800 & 0.6794 & 0.6442 & 0.6122 & 0.6192 & 0.3167 & 0.2746 & 0.2812 & 0.5467 & 0.5223 & 0.5266 \\
& GPT-4o~\cite{openai2024gpt4o}                      & 0.8239 & 0.8333 & 0.8263 & 0.6143 & 0.5778 & 0.5864 & 0.2807 & 0.2304 & 0.2407 & 0.5729 & 0.5471 & 0.5511 \\
& UNOGrasp~\cite{jiao2026obstruction}                 & 0.9150 & 0.9150 & 0.9150 & \textbf{0.7333} & \textbf{0.6828} & \textbf{0.6989} & \underline{0.5825} & \underline{0.4997} & \underline{0.5238} & \textbf{0.7436} & \underline{0.6992} & \underline{0.7126} \\
& InternVL3.5~\cite{wang2025internvl3}                & 0.3024 & 0.6100 & 0.3704 & 0.6163 & 0.6356 & 0.6045 & 0.3180 & 0.2935 & 0.2856 & 0.4122 & 0.5130 & 0.4201 \\
% \midrule
& \textbf{Our CPOR-Grasp}                                      & \textbf{0.9383} & \textbf{0.9383} & \textbf{0.9383} & \underline{0.6861} & \underline{0.6776} & \underline{0.6724} & \textbf{0.5861} & \textbf{0.5525} & \textbf{0.5559} & \underline{0.7368} & \textbf{0.7228} & \textbf{0.7222} \\

% \midrule
% \multicolumn{13}{l}{\textbf{B. Real-world test set}}\\
\midrule
\multirow{6}{*}{\rotatebox{90}{\textbf{B. Real-world}}} & Gemini Robotics-ER-1.6~\cite{gemini_robotics_er_16} & \underline{0.8828} & \underline{0.8867} & \underline{0.8839} & 0.6517 & 0.6156 & 0.6234 & 0.4188 & 0.3697 & 0.3758 & 0.6682 & 0.6428 & 0.6463 \\
& Qwen3.5~\cite{qwen35}                              & 0.6167 & 0.6267 & 0.6200 & \underline{0.7273} & 0.6889 & \underline{0.6951} & 0.2668 & 0.2749 & 0.2531 & 0.5569 & 0.5490 & 0.5426 \\
& GPT-4o~\cite{openai2024gpt4o}                      & 0.7283 & 0.7367 & 0.7311 & 0.6472 & 0.6033 & 0.6148 & 0.2290 & 0.1793 & 0.1912 & 0.5574 & 0.5306 & 0.5361 \\
& UNOGrasp~\cite{jiao2026obstruction}                 & 0.8500 & 0.8500 & 0.8500 & \textbf{0.8017} & \textbf{0.7356} & \textbf{0.7556} & \textbf{0.4916} & \underline{0.4051} & \underline{0.4318} & \textbf{0.7309} & \underline{0.6826} & \underline{0.6974} \\
& InternVL3.5~\cite{wang2025internvl3}             & 0.5708 & 0.5833 & 0.5747 & 0.6750 & 0.6272 & 0.6367 & 0.2627 & 0.2398 & 0.2369 & 0.5205 & 0.5014 & 0.5009 \\
% \midrule
& \textbf{Our CPOR-Grasp}                                     & \textbf{0.9017} & \textbf{0.9017} & \textbf{0.9017} & 0.7206 & \underline{0.6883} & \underline{0.6951} & \underline{0.4885} & \textbf{0.4423} & \textbf{0.4468} & \underline{0.7195} & \textbf{0.6948} & \textbf{0.6985} \\
\bottomrule
\end{tabular*}
\vspace{-1.5mm}
\end{table*}

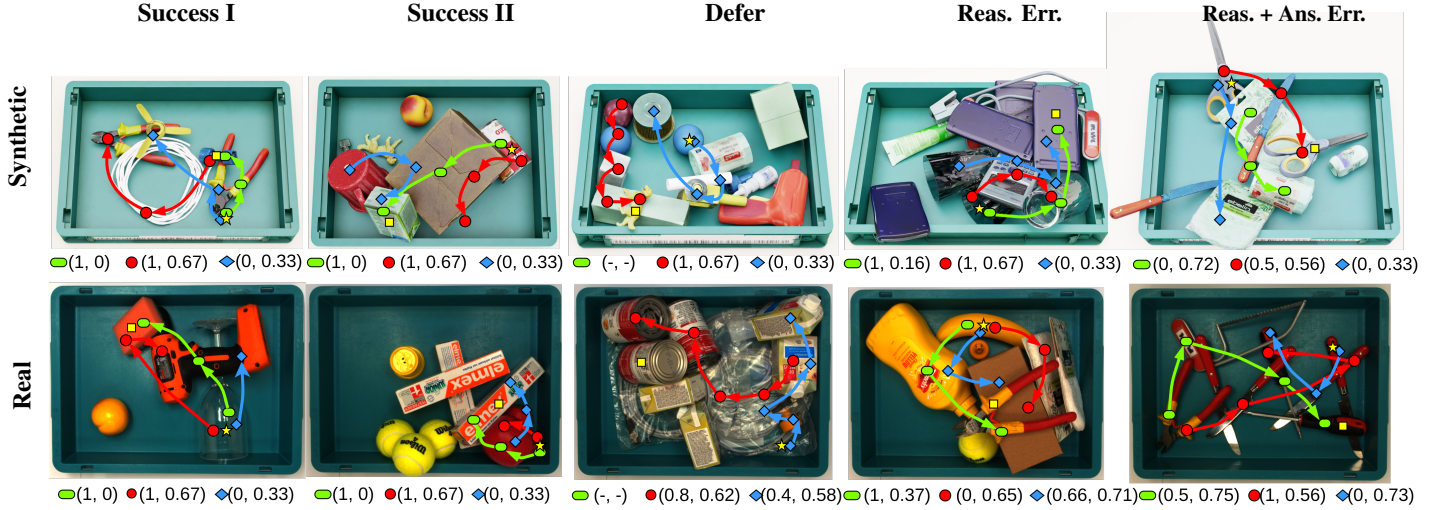
\begin{figure*}[!t]     
    \centering     
    \begin{tikzpicture}          
        \node[inner sep=0, anchor=south west] (img) at (0,0) {%             
            \includegraphics[width=\textwidth]             
            {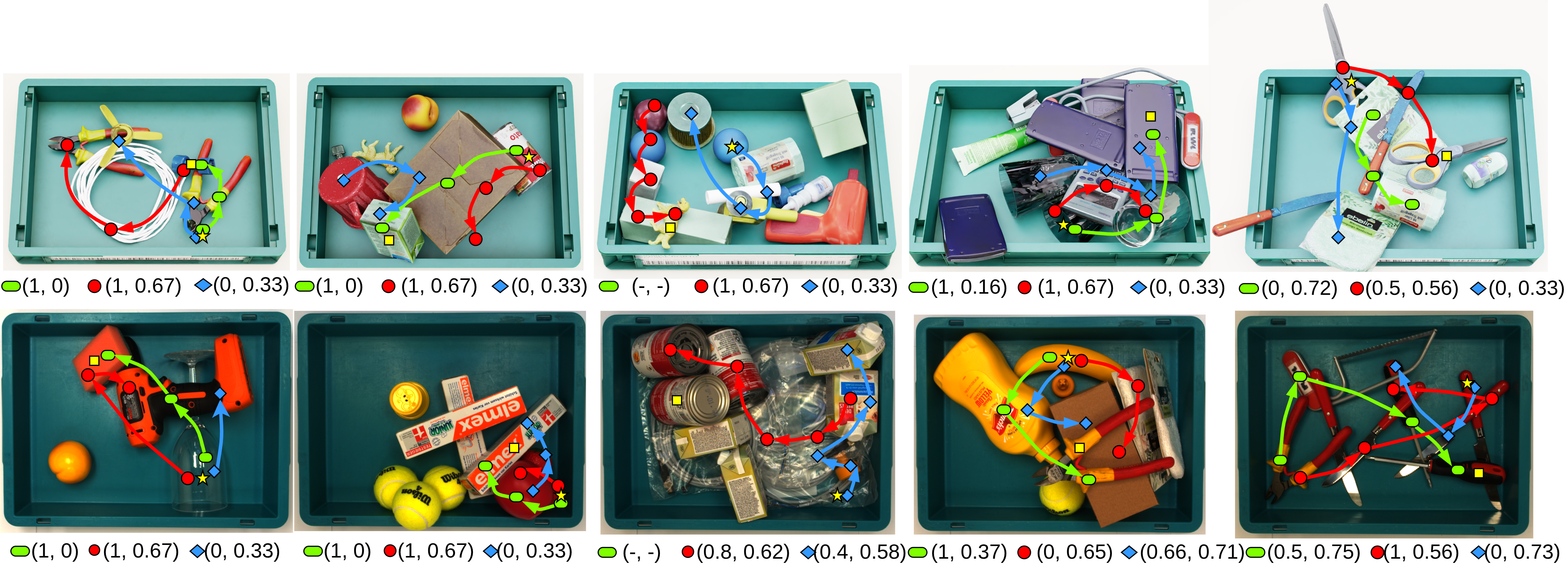}};          
        
        % ============================================================         
        % Column labels -- Đã phóng to font chữ (9.5pt / 10.0pt)
        % ============================================================         
        \node[             
            anchor=south,             
            font=\bfseries\fontsize{9.5}{10.0}\selectfont 
        ] at ($(img.north west)!0.10!(img.north east)+(0,-2.5mm)$)         
        {Success I};          
        
        \node[             
            anchor=south,             
            font=\bfseries\fontsize{9.5}{10.0}\selectfont 
        ] at ($(img.north west)!0.30!(img.north east)+(0,-2.5mm)$)         
        {Success II};          
        
        \node[             
            anchor=south,             
            font=\bfseries\fontsize{9.5}{10.0}\selectfont 
        ] at ($(img.north west)!0.50!(img.north east)+(0,-2.5mm)$)         
        {Defer};          
        
        \node[             
            anchor=south,             
            font=\bfseries\fontsize{9.0}{9.5}\selectfont 
        ] at ($(img.north west)!0.70!(img.north east)+(0,-2.5mm)$)         
        {Reas. Err.};          
        
        \node[             
            anchor=south,             
            font=\bfseries\fontsize{8.5}{9.0}\selectfont 
        ] at ($(img.north west)!0.90!(img.north east)+(0,-2.5mm)$)         
        {Reas. + Ans. Err.};          
        
        % ============================================================         
        % Row labels -- Đã phóng to font chữ và đẩy dịch ra ngoài một chút
        % ============================================================         
        \node[             
            rotate=90,             
            anchor=south,             
            font=\bfseries\fontsize{9.5}{10.0}\selectfont 
        ] at ($(img.south west)!0.75!(img.north west)+(-1.2mm,0)$)         
        {Synthetic};          
        
        \node[             
            rotate=90,             
            anchor=south,             
            font=\bfseries\fontsize{9.5}{10.0}\selectfont 
        ] at ($(img.south west)!0.25!(img.north west)+(-1.2mm,0)$)         
        {Real};      
    \end{tikzpicture}      
    \vspace{-1.6mm}      
    \caption{Qualitative obstruction reasoning on synthetic (top) and real-world (bottom) UNOBench scenes. \targetstar\ marks the target and \topobstructor\ the top obstructor. Reasoning traces are shown for \unograspmarker\ UNOGrasp~\cite{jiao2026obstruction}, \bluediamond\ Gemini Robotics-ER 1.6, and \greenroundbox\ CPOR-Grasp. (SR-F1/MP-NED) scores are reported at the bottom.}     
    \label{fig:qualitative} 
\end{figure*}

Tables~\ref{tab:path-bench} and \ref{tab:obj-bench} show that CPOR-Grasp's gains are concentrated in full removal-chain reasoning and difficult scenes. It achieves the best path-level performance across all difficulty levels on both test sets, reaching SR-$F_1$ scores of 0.5491 and 0.4583 on Hard scenes, compared with 0.3276 and 0.2297 for UNOGrasp, while reducing overall MP-NED to 0.2247 and 0.2423. Improvements at the object level are smaller: UNOGrasp remains strongest on Medium scenes, and overall real-world OR-$F_1$ is nearly tied (0.6974 vs. 0.6985). These results suggest that a correct next action can be predicted despite incomplete removal-chain reasoning (Fig.~\ref{fig:qualitative}).

% \begin{figure}[!t]
% \centering
% \includegraphics[width=0.47\textwidth]{figures/qualitative_syn_real.png}
% \caption{Qualitative obstruction reasoning on synthetic (top) and real-world (bottom) UNOBench scenes. \targetstar\ marks the target object and \topobstructor\ marks the top obstructor. The reasoning traces are denoted by \unograspmarker\ UNOGrasp~\cite{jiao2026obstruction}, \bluediamond\ Gemini Robotics-ER 1.6, \greenroundbox\ CPOR-Grasp.}
% \label{fig:qualitative}
% \end{figure}

% Figure~\ref{fig:qualitative} illustrates these cases: a correct next object can follow an incorrect intermediate chain, while the cyclic and defer examples correspond to the DAG and selective-decision analyses below. The joint failure shows the remaining limitation when upstream evidence is incorrect.

\subsection{Reliability and Ablation Analysis}
\label{sec:reliability}

\begin{figure}[!t]
\centering
\includegraphics[width=0.47\textwidth]{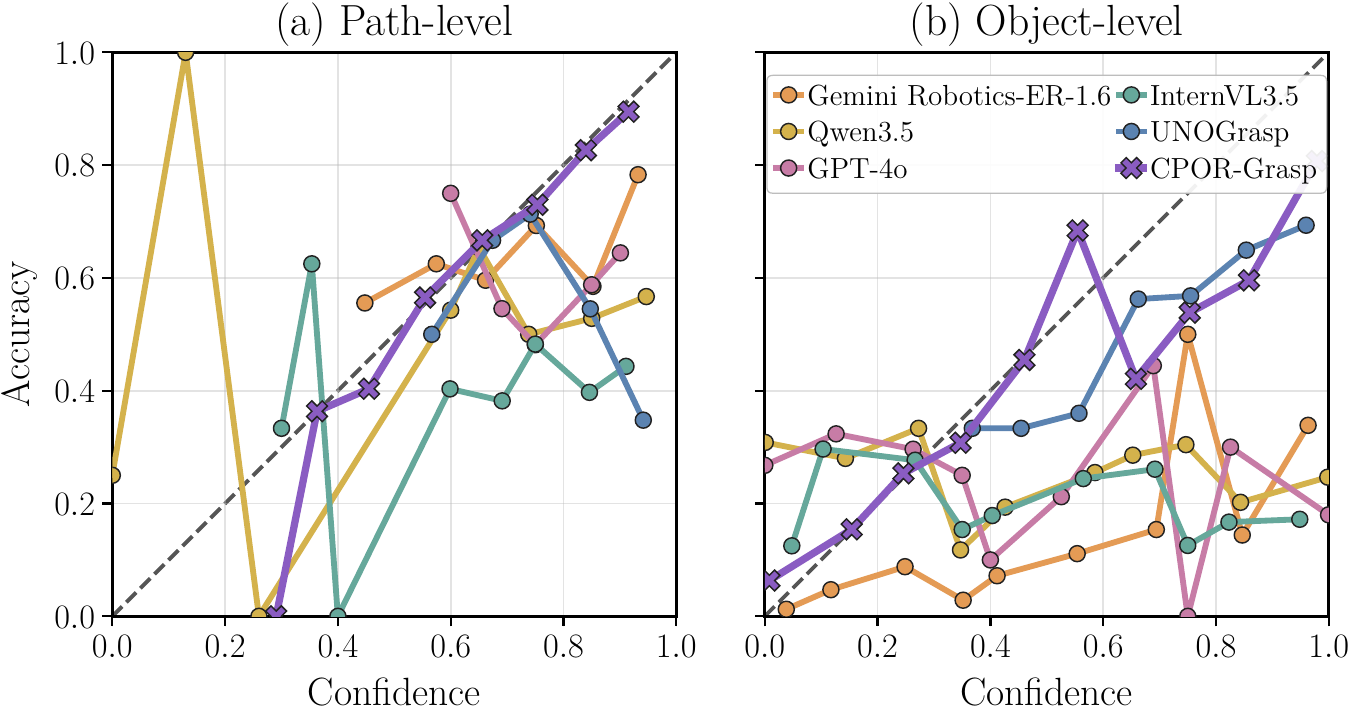}
\caption{Reliability diagrams for relation (a) and object (b) confidence. Bins match the ECE computation. CPOR-Grasp is closest to the ideal $y=x$ overall; general-purpose VLMs use the common auxiliary confidence extractor, while UNOGrasp keeps its released scoring semantics.}
\label{fig:reliability}
\end{figure}

Geometry alone (UOAIS-FT, Table~\ref{tab:geometry}) already reaches 0.5711 SR-$F_1$ on the synthetic set, above every RGB baseline. Fusion adds 0.046 SR-$F_1$ on top of it, and removal-order conditioning adds 0.014 OR-$F_1$ (0.7083 to 0.7222, Table~\ref{tab:cpor-ablation}).  Table~\ref{tab:reliability-baselines} evaluates the probabilities used by structured inference. UNOGrasp keeps its released confidence semantics. We report ECE, Brier score, NLL, and AUROC~\cite{guo2017calibration}. UNOGrasp has lower relation ECE than Gemini Robotics-ER-1.6 (0.1088 against 0.1416), but relation AUROC of 0.4125 is below chance. CPOR-Grasp improves both, reaching relation and object ECE of 0.0185 and 0.0824 with AUROC of 0.6554 and 0.8624. Fig.~\ref{fig:reliability} shows residual over-confidence at high object confidence.
In Table~\ref{tab:prob-ablation}, Platt scaling fits one scene-independent affine map of the log-odds, and temperature scaling fits only its slope. Temperature scaling alone has lower relation ECE than adaptive Platt (0.0278 against 0.0478). Although temperature scaling yields lower standalone relation ECE, adaptive Platt composes best with reliability-gated fusion, yielding the lowest final ECE and NLL and the highest AUROC.
% We choose adaptive Platt because it composes with gated fusion to the best overall reliability, ECE 0.0185. Its regularization $\lambda$ is chosen inside the fitting split by scene-grouped cross-validation on NLL. 
% Naive fusion adds the two log-odds with unit weights and no prior term. It worsens Brier and AUROC to 0.2334 and 0.5537, whereas gated fusion improves them to 0.1848 and 0.6554.

% Table~\ref{tab:prob-ablation} separates calibration from fusion. Temperature scaling reduces relation ECE to 0.0278, with AUROC at 0.5961. Naive fusion worsens Brier score and AUROC to 0.2334 and 0.5537, whereas reliability-gated fusion reaches 0.0185 and 0.6554. Downstream object ECE decreases from 0.4718 to 0.1357, 0.1084, and 0.0824 across successive stages, while AUROC increases from 0.6725 to 0.7733, 0.8413, and 0.8624, showing that the reliability gains propagate to the action marginals.

\begin{table}[!t]
\centering
\caption{Ablation study on 3D Geometry information extraction.}
\label{tab:geometry}
\vspace{-2.5mm}
\scriptsize 
\setlength{\tabcolsep}{0pt} 
\renewcommand{\arraystretch}{1.0}

\begin{tabular*}{\columnwidth}{@{\extracolsep{\fill}}l ccc ccc@{}}
\toprule
& \multicolumn{3}{c}{Synthetic} & \multicolumn{3}{c}{Real-world} \\
\cmidrule(lr){2-4}\cmidrule(lr){5-7}
Method & SR-$F_1$$\uparrow$ & OR-$F_1$$\uparrow$ & MP-NED$\downarrow$ & SR-$F_1$$\uparrow$ & OR-$F_1$$\uparrow$ & MP-NED$\downarrow$ \\
\midrule
UOAIS~\cite{back2022unseen} & \underline{0.5403} & \underline{0.6435} & 0.2776 & 0.2796 & 0.3607 & 0.4234 \\
D3G~\cite{rabino2025modern}   & 0.4362 & 0.6321 & \underline{0.2732} & \underline{0.3207} & \underline{0.5494} & \underline{0.3099} \\
\textbf{UOAIS-FT}    & \textbf{0.5711} & \textbf{0.6920} & \textbf{0.2378} & \textbf{0.5084} & \textbf{0.6282} & \textbf{0.2649} \\
\bottomrule
\end{tabular*}
\vspace{-1.5mm}
\end{table}

\begin{table}[!t]
\centering
\caption{ Probability reliability on the UNOBench synthetic test set.}
\label{tab:reliability-baselines}
\vspace{-2.5mm}

\scriptsize 
\setlength{\tabcolsep}{0pt} 
\renewcommand{\arraystretch}{1.0}

\begin{tabular*}{\columnwidth}{@{\extracolsep{\fill}}l l c c c c@{}}
\toprule
Level & Method & ECE$\downarrow$ & Brier$\downarrow$ & NLL$\downarrow$ & AUROC$\uparrow$ \\
\midrule
\multirow{5}{*}{Relation} & Gemini Robotics-ER-1.6~\cite{gemini_robotics_er_16} & 0.1416 & \underline{0.2149} & \underline{0.6777} & \underline{0.5985} \\
                          & Qwen3.5~\cite{qwen35}                & 0.2484 & 0.3252 & 0.9415 & 0.4992 \\
                          & GPT-4o~\cite{openai2024gpt4o}                 & 0.2568 & 0.3056 & 0.8562 & 0.5291 \\
                          & UNOGrasp~\cite{jiao2026obstruction}               & \underline{0.1088} & 0.2570 & 0.7564 & 0.4125 \\
                          & InternVL3.5~\cite{wang2025internvl3}     & 0.3135 & 0.3460 & 0.9310 & 0.5167 \\
                          & \textbf{CPOR-Grasp}          & \textbf{0.0185} & \textbf{0.1848} & \textbf{0.5533} & \textbf{0.6554} \\
\midrule
\multirow{5}{*}{Object}   & Gemini Robotics-ER-1.6~\cite{gemini_robotics_er_16} & 0.4718 & 0.4421 & 2.9000 & 0.6725 \\
                          & Qwen3.5~\cite{qwen35}                & 0.5781 & 0.5759 & 7.4644 & 0.4698 \\
                          & GPT-4o~\cite{openai2024gpt4o}                 & 0.6504 & 0.6505 & 8.8984 & 0.4425 \\
                          & UNOGrasp~\cite{jiao2026obstruction}               & \underline{0.2289} & \underline{0.2765} & \underline{0.9650} & \underline{0.5843} \\
                          & InternVL3.5~\cite{wang2025internvl3}    & 0.5655 & 0.5217 & 2.5109 & 0.4536 \\
                          & \textbf{CPOR-Grasp}          & \textbf{0.0824} & \textbf{0.1028} & \textbf{0.9487} & \textbf{0.8624} \\
\bottomrule
\end{tabular*}
\vspace{-1.5mm}
\end{table}

\begin{table}[!t]
\centering
\caption{Reliability ablation on the UNOBench synthetic test set. PM denotes posterior marginalization.}
\label{tab:prob-ablation}
\vspace{-2.0mm}

\scriptsize
\setlength{\tabcolsep}{3.5pt}
\renewcommand{\arraystretch}{1.0}

\begin{tabular*}{\columnwidth}{@{\extracolsep{\fill}}l l c c c c@{}}
\toprule
\textbf{Level} & \textbf{Configuration} & \textbf{ECE}$\downarrow$ & \textbf{Brier}$\downarrow$ & \textbf{NLL}$\downarrow$ & \textbf{AUROC}$\uparrow$ \\
\midrule

\multirow{7}{*}{Relation}
& Gemini Robotics-ER-1.6~\cite{gemini_robotics_er_16}    & 0.1416 & 0.2149 & 0.6777 & 0.5985 \\
& Temp scaling~\cite{guo2017calibration}      & \underline{0.0278} & 0.1941 & 0.5759 & 0.5962 \\
& Platt scaling~\cite{platt1999probabilistic}            & 0.0619 & 0.1979 & 0.5837 & 0.5985 \\
& Isotonic~\cite{zadrozny2002transforming}               & 0.0453 & 0.1963 & 0.6347 & 0.6033 \\
& Adaptive Platt               & 0.0478 & 0.1978 & 0.5823 & 0.6084 \\
& Temp scaling + Fusion & 0.0734 & 0.1915 & 0.5721 & \underline{0.6494} \\
& Platt scaling + Fusion & 0.0435 & 0.1886 & \underline{0.5649} & 0.6085 \\
& Isotonic + Fusion & 0.0520 & \underline{0.1869} & 0.6022 & 0.6311 \\
& Adaptive Platt + Naive fusion & 0.1401 & 0.2334 & 0.6625 & 0.5537 \\
& Adaptive Platt + Fusion & \textbf{0.0185} & \textbf{0.1848} & \textbf{0.5533} & \textbf{0.6554} \\

\midrule

\multirow{4}{*}{Object}
& Gemini Robotics-ER-1.6~\cite{gemini_robotics_er_16}  & 0.4718 & 0.4421 & 2.9000 & 0.6725 \\
& Adaptive Platt + PM               & 0.1357 & 0.1413 & 1.7269 & 0.7733 \\
& Adaptive Platt + Fusion + PM      & \underline{0.1084} & \underline{0.1188} & \underline{1.0612} & \underline{0.8413} \\
& Adaptive Platt + Fusion + DAG + PM & \textbf{0.0824} & \textbf{0.1028} & \textbf{0.9487} & \textbf{0.8624} \\

\bottomrule
\end{tabular*}
\vspace{-2.0mm}
\end{table}

\subsection{Structured Inference and Selective Decisions}
\label{sec:structured-selective}
Table~\ref{tab:cpor-ablation} isolates removal-order (DAG) conditioning, the restriction to graphs that admit one. Rel-ECE and O-ECE denote relation and object ECE. Panel A shows cumulative gains, and removal-order conditioning moves only OR-$F_1$. Panel B localizes this effect. All 1660 $\mu=0$ scenes are unchanged. On the 140 $\mu>0$ scenes, OR-$F_1$ rises from 0.4339 to 0.6127 and O-ECE falls from 0.3805 to 0.1150.  Conflict grows with difficulty (Table~\ref{tab:mu}), from 1.3\% of Easy to 15.0\% of Hard scenes and 7.8\% overall. When present, the median excluded mass is high (0.8013 overall), indicating substantial conflict with acyclicity.

% \textcolor{blue}{NOTE: there is more than one structural error - not only cycle as reported. Inconsistent depth, reachability,... are also structural errors caused by VLMs' structural hallucination. Cycle required DAG to solve so we focus on this more than the others}
% Table~\ref{tab:cpor-ablation} evaluates how the preceding probability improvements affect task decisions and isolates DAG conditioning. Here, $\mu=1-Z$ is the product-model mass outside DAG support; it measures cyclic off-DAG support only, not other structural or perception errors.

\begin{table}[!t]
\centering
\caption{CPOR-Grasp mechanism ablation on the synthetic test set.}
\label{tab:cpor-ablation}
\vspace{-2.0mm}

\scriptsize
\renewcommand{\arraystretch}{1.0}

% =========================================================
% Panel A
% =========================================================
\setlength{\tabcolsep}{4.0pt}
\begin{tabular*}{\columnwidth}{@{\extracolsep{\fill}}l c c c@{}}
\toprule
\multicolumn{4}{l}{A. Cumulative CPOR-Grasp ablation \hfill \textit{$N=1800$ scenes}}\\
\midrule
\textbf{Configuration} & \textbf{SR-$F_1$}$^{\dagger}\uparrow$ & \textbf{OR-F1}$\uparrow$ & \textbf{MP-NED}$^{\dagger}\downarrow$ \\
\midrule
Gemini Robotics-ER-1.6~\cite{gemini_robotics_er_16}                         & 0.4705 & 0.6767 & 0.3036 \\
Adaptive Platt + PM                & 0.5668 & 0.6814 & 0.2523 \\
Adaptive Platt + Fusion + PM       & \textbf{0.6174} & 0.7083 & \textbf{0.2247} \\
Adaptive Platt + Fusion + DAG + PM & \textbf{0.6174} & \textbf{0.7222} & \textbf{0.2247} \\
\end{tabular*}
{\scriptsize\raggedright
$^{\dagger}$SR-$F_1$ and MP-NED are measured before DAG conditioning and are therefore unchanged.\par}
\vspace{1.5mm}

% =========================================================
% Panel B
% =========================================================
\setlength{\tabcolsep}{1.0pt}
\begin{tabular*}{\columnwidth}{@{\extracolsep{\fill}}l l c c c@{}}
\toprule
\multicolumn{5}{l}{B. Conflict-conditioned DAG effect}\\
\midrule
\textbf{Condition} & \textbf{Configuration} & \textbf{OR-F1}$\uparrow$ & \textbf{O-ECE}$\downarrow$ & \textbf{O-AUROC}$\uparrow$ \\
\midrule
\multirow{2}{*}{\shortstack{$\mu=0$\\($N=1660$)}} & Adaptive Platt + Fusion + PM       & 0.7314 & 0.0824 & 0.8609 \\
                                                   & Adaptive Platt + Fusion + DAG + PM & 0.7314 & 0.0824 & 0.8609 \\
\addlinespace[0.5mm]
\multirow{2}{*}{\shortstack{$\mu>0$\\($N=140$)}}  & Adaptive Platt + Fusion + PM       & 0.4339 & 0.3805 & 0.7712 \\
                                                   & Adaptive Platt + Fusion + DAG + PM & \textbf{0.6127} & \textbf{0.1150} & \textbf{0.8243} \\
\bottomrule
\end{tabular*}
\vspace{-2.0mm}
\end{table}

\begin{table}[!t]
\centering
\caption{Product-model mass outside graph support on the synthetic test set.}
\label{tab:mu}
\vspace{-1.0mm}
\scriptsize
\setlength{\tabcolsep}{0.5pt}
\renewcommand{\arraystretch}{1.03}
\begin{tabular*}{\columnwidth}{@{\extracolsep{\fill}}l c c c c@{}}
\toprule
\textbf{Tier} & \textbf{$N$} & \textbf{$\mu>0$} & \textbf{Mean $\mu$} & \textbf{Median [IQR] $\mu\mid\mu>0$} \\
\midrule
Easy & 600 & 8 (1.3\%) & 0.0096 & 0.7494 [0.6417, 0.8518] \\
Medium & 600 & 42 (7.0\%) & 0.0518 & 0.7611 [0.6914, 0.8235] \\
Hard & 600 & 90 (15.0\%) & 0.1183 & 0.8179 [0.7246, 0.8855] \\
Overall & 1800 & 140 (7.8\%) & 0.0599 & 0.8013 [0.6955, 0.8704] \\
\bottomrule
\end{tabular*}
\vspace{-1.0mm}
\end{table}

\noindent\textit{Marginalization beyond feasibility.}
RSR is the retrieval-step success rate, the fraction of correct next actions. 
With fixed upstream probabilities and $N=1800$, $\Delta\mathrm{RSR}=(b-c)/N$, where $b$ and $c$ denote paired corrections and regressions, respectively. The 95\% CIs are obtained by paired bootstrap over scenes. No-DAG marginalizes under the unconditioned product model \eqref{eq:weight}, and Single-MAP is the $K=1$. Table~\ref{tab:paired} gives 90 corrections against 23 regressions for No-DAG, and 32 against 11 for Single-MAP, with both intervals excluding zero. All 113 decision changes in the first comparison occur within the 140 scenes with $\mu>0$, because removal-order conditioning is the identity elsewhere. One feasible graph is therefore no substitute for marginalizing over several.

\begin{table}[!b]
\centering
\caption{Paired next-action comparison under fixed %upstream 
probabilities.}
\label{tab:paired}

\vspace{-2.0mm}
\scriptsize
\setlength{\tabcolsep}{0.0pt}
\renewcommand{\arraystretch}{0.8}

\begin{tabular*}{\columnwidth}{
@{\extracolsep{\fill}}
l c c c c
@{}
}
\toprule
\textbf{Comparison}
& $\boldsymbol{b}$
& $\boldsymbol{c}$
& $\boldsymbol{\Delta}$\textbf{RSR}
& \textbf{95\% CI} \\
\midrule
No-DAG $\rightarrow$ DAG marginal
& 90 & 23 & +3.7 pp & [+2.57,+4.88] \\

Single-MAP $\rightarrow$ DAG marginal
& 32 & 11 & +1.2 pp & [+0.42,+1.87] \\
\bottomrule
\end{tabular*}

\vspace{-1.0mm}
\end{table}

\noindent\textit{Certified Top-$K$.}
A scene is eligible when fusion yields at least one candidate pair. This holds for 1553 of the 1800 records, all below the exact-enumeration cutoff of 20 candidate pairs. The adaptive rule uses $K_{\max}=256$ and $\epsilon_{\mathrm{tol}}=0.05$. $\dTV(\PD,\PK)=1-Z_K/Z$ is the truncated posterior mass, exact agreement compares the selected action with exact inference, and certification applies the sufficient test of Corollary~\ref{cor:stability}. The exact-enumeration row of Table~\ref{tab:topk} reads 0/0, 100\%, and a dash by definition. Adaptive Top-$K$ matches fixed $K=100$ at 99.74\% exact agreement while retaining 22.68 graphs on average instead of 51.63, against 1281.53 under exact enumeration. In wall clock, the adaptive rule averages 0.134\,ms per scene against 20.9\,ms for exact enumeration. Its worst case stays at 8.5\,ms against 8.9\,s. Certification at 92.98\% below 99.74\% agreement is expected, because the stability test is sufficient but not necessary. Certified means agreement with exact inference under the model, not correctness.

\begin{table}[!t]
\centering
\caption{Top-$K$ approximation and adaptive certification.}
\label{tab:topk}
\vspace{-2.0mm}
\scriptsize
\setlength{\tabcolsep}{0.0pt}
\renewcommand{\arraystretch}{1.00}
\begin{tabular*}{\columnwidth}{@{\extracolsep{\fill}}l c c c c c@{}}
\toprule
& \multicolumn{2}{c}{\textbf{Approximation validity}} & \multicolumn{3}{c}{\textbf{Computational utility}} \\
\cmidrule(lr){2-3}\cmidrule(lr){4-6}
Method & Mean/Max $d_{\rm TV}$ & Exact agree. & Cert. rate & Mean $K$ & Mean/Max \textit{(10 runs - ms)} \\
\midrule
\textit{reference} & \TBD/\TBD & \TBD & \TBD & 1281.53 & 20.910$\pm$0.180/8912$\pm$41 \\
$K{=}20$ & 0.093/0.963 & 99.19\% & 86.28\% & \textbf{13.47} & 0.066$\pm$0.001/0.098$\pm$0.002 \\
$K{=}50$ & 0.059/0.935 & \underline{99.54\%} & 90.53\% & 27.84 & 0.148$\pm$0.002/0.274$\pm$0.004 \\
$K{=}100$ & \underline{0.041/0.89} & \textbf{99.74\%} & \underline{92.45\%} & 51.63 & 0.297$\pm$0.003/0.586$\pm$0.006 \\
Ours & \textbf{0.027/0.500} & \textbf{99.74\%} & \textbf{92.98\%} & \underline{22.68} & 0.134$\pm$0.002/8.471$\pm$0.078 \\
\bottomrule
\end{tabular*}
\vspace{-1.0mm}
\end{table}

% \begin{table}[!t]
% \centering
% \caption{Top-$K$ approximation and adaptive certification. 
% % Approximation validity is evaluated on the exact-evaluable subset; mode count and certification use all eligible scenes ($N=1553$).
% }
% \label{tab:topk}
% \vspace{-1.0mm}
% \scriptsize
% \setlength{\tabcolsep}{1.0pt}
% \renewcommand{\arraystretch}{1.03}
% \begin{tabular*}{\columnwidth}{@{\extracolsep{\fill}}l c c c c@{}}
% \toprule
% & \multicolumn{2}{c}{\textbf{Approximation validity}} & \multicolumn{2}{c}{\textbf{Computational utility}} \\
% \cmidrule(lr){2-3}\cmidrule(lr){4-5}
% \textbf{Method} & \textbf{Mean/Max $d_{\rm TV}$} & \textbf{Exact agree.} & \textbf{Cert. rate} & \textbf{Mean $K$} \\
% \midrule
% \textit{Exact enumeration (ref)} & \TBD/\TBD & \TBD & \TBD & 1281.53 \\
% Fixed Top-$K$ ($K{=}20$) & 0.0931/0.9628 & 99.19\% & 86.28\% & \textbf{13.47} \\
% Fixed Top-$K$ ($K{=}50$) & 0.0593/0.9346 & \underline{99.54\%} & 90.53\% & 27.84  \\
% Fixed Top-$K$ ($K{=}100$) & \underline{0.04103/0.8913} & \textbf{99.74\%} & \underline{92.45\%} & 51.63 \\
% Adaptive certified (Ours) & \textbf{0.02652/0.4999} & \textbf{99.74\%} & \textbf{92.98\%} & \underline{22.68} \\
% \bottomrule
% \end{tabular*}
% \vspace{-1.0mm}
% \end{table}

\begin{figure}[!t]
\centering
\includegraphics[width=0.47\textwidth]{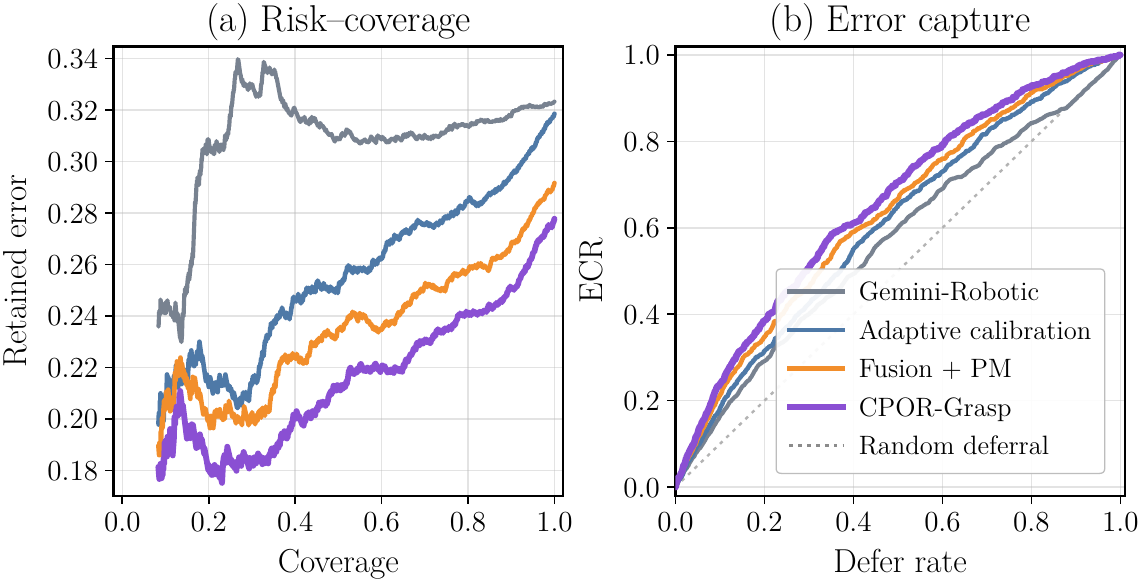}
\caption{Selective decision behavior. (a) Retained task error versus coverage; (b) Error Capture Rate versus defer rate $D$.}
\label{fig:selective}
\end{figure}

\noindent\textit{Selective execution.}
Forced decisions set $\tau=0$ in \eqref{eq:policy} with no deferral, and the sweep in Fig.~\ref{fig:selective} varies $\tau$. At defer rate $D$, coverage is $C=1-D$ and retained error is $\rho(D)\downarrow=1-\mathrm{OR}$-$F_1(D)$ on accepted scenes. $\mathrm{AURC}_{F_1}$ is the area under this risk-coverage curve~\cite{geifman2019bias}. The error capture rate $\mathrm{ECR}(D)=1-(1-D)\rho(D)/\rho(0)$ is the fraction of forced-decision error removed by deferring, and random deferral gives $\mathrm{ECR}=D$~\cite{geifman2017selective}. Calibration barely changes forced OR-$F_1$ but lowers $\mathrm{AURC}_{F_1}$, and fusion and removal-order conditioning lower it further. CPOR-Grasp also has the highest ECR.

\subsection{Cross-VLM Generalization and Real-Robot Evaluation}
\label{sec:cross-vlm}
\begin{figure}[t]
    \centering
    % Dùng TikZ để phủ một lớp lọc giúp ảnh đậm màu và bớt chói sáng
    \begin{tikzpicture}
        \node[inner sep=0, anchor=south west] (img) at (0,0) {%
            \includegraphics[width=0.9\columnwidth]{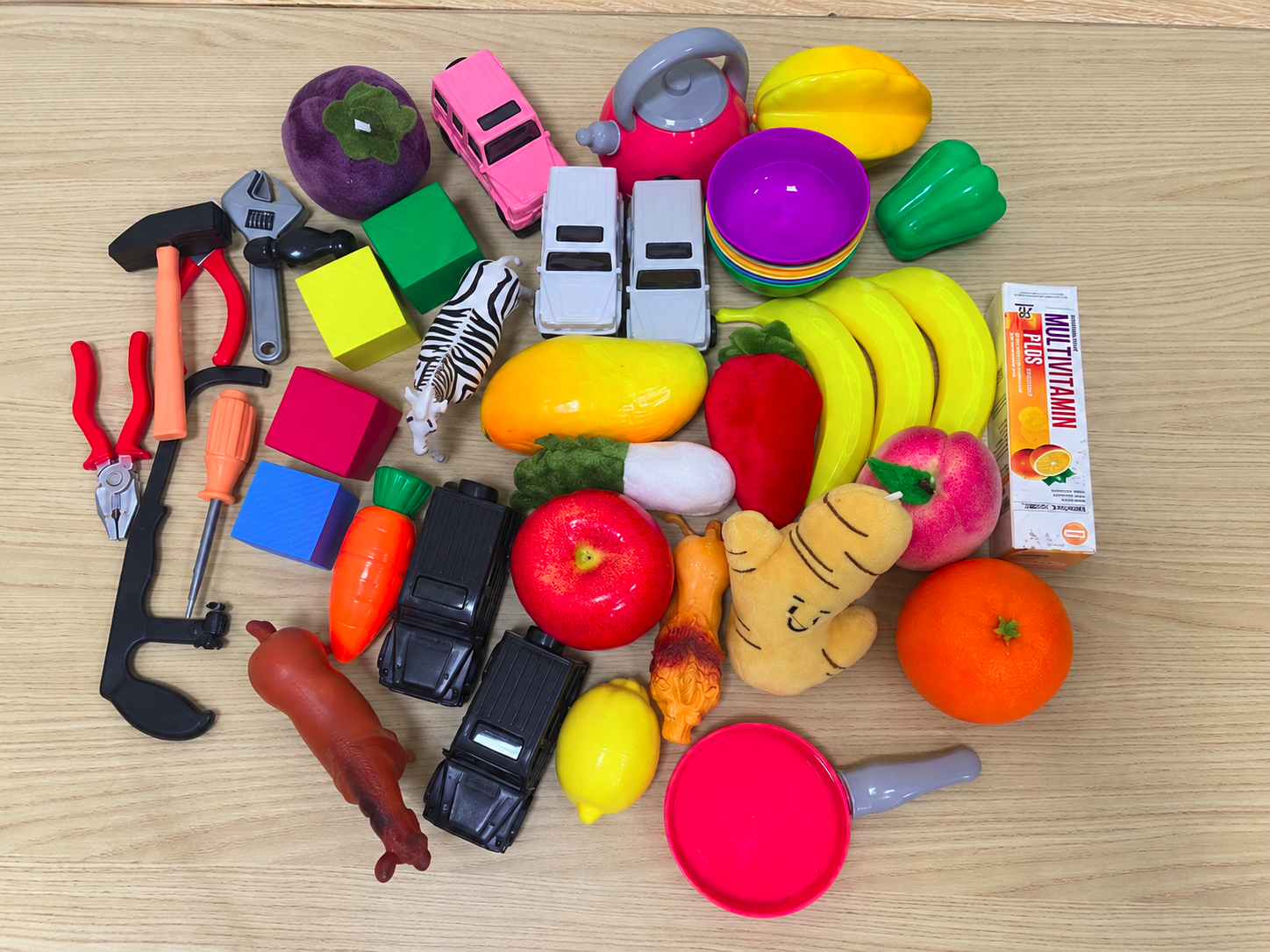}%
        };
        \path[fill=black, opacity=0.1, blend mode=multiply] (img.south west) rectangle (img.north east);
    \end{tikzpicture}
    \caption{Physical object set used for the real-robot evaluation. The set contains 40 object instances drawn from diverse object categories, including household tools, toy vehicles, containers, produce, geometric blocks, and other everyday items.}
    \label{fig:robot-objects}
\end{figure}

%Each VLM is evaluated raw and after the same pipeline. Nothing is refit per backbone. The calibrator, the fusion weights, and the decoding threshold stay fixed, and only the candidate-pool base rate entering the fusion intercept is recomputed. All five backbones improve in SR-$F_1$ and OR-$F_1$ and reduce both ECE values. For Qwen3.5, InternVL3.5, and Gemma3, fused OR-$F_1$ stays below the geometry-only branch (0.6920, Table~\ref{tab:geometry}), so geometry carries the gain for weaker backbones.
To test the backbone generality, each VLM is evaluated raw and after the same pipeline. The calibrator, the fusion weights, and the decoding threshold stay fixed, and only the candidate-pool base rate entering the fusion intercept is recomputed. All five backbones improve in SR-$F_1$ and OR-$F_1$ and reduce both ECE values. For Qwen3.5, InternVL3.5, and Gemma3, fused OR-$F_1$ stays below the geometry-only branch (0.6920, Table~\ref{tab:geometry}), so geometry carries the gain for weaker backbones.
\begin{table}[t]
\centering
\caption{Cross-VLM performance and reliability on the UNOBench synthetic test set.}
\label{tab:cross-vlm}
\vspace{-2.0mm}
\scriptsize
\setlength{\tabcolsep}{2.6pt}
\renewcommand{\arraystretch}{1.00}
\resizebox{\columnwidth}{!}{%
\begin{tabular}{@{}lcccc@{}}
\toprule
\textbf{VLM} & \textbf{SR-$F_1$}$\uparrow$ & \textbf{OR-$F_1$}$\uparrow$ & \textbf{Rel-ECE}$\downarrow$ & \textbf{O-ECE}$\downarrow$ \\
\midrule
Gemini Robotics-ER-1.6~\cite{gemini_robotics_er_16} & 0.4705 $\rightarrow$ \textbf{0.6174} & 0.6767 $\rightarrow$ \textbf{0.7222} & 0.1416 $\rightarrow$ \textbf{0.0185} & 0.4718 $\rightarrow$ \textbf{0.0824} \\
GPT-4o~\cite{openai2024gpt4o} & 0.3817 $\rightarrow$ \textbf{0.6100} & 0.5511 $\rightarrow$ \textbf{0.6930} & 0.2568 $\rightarrow$ \textbf{0.1417} & 0.6504 $\rightarrow$ \textbf{0.1046} \\
Qwen3.5~\cite{qwen35} & 0.4335 $\rightarrow$ \textbf{0.6129} & 0.5266 $\rightarrow$ \textbf{0.6521} & 0.2484 $\rightarrow$ \textbf{0.1503} & 0.5781 $\rightarrow$ \textbf{0.1399} \\
InternVL3.5~\cite{wang2025internvl3} & 0.3549 $\rightarrow$ \textbf{0.5696} & 0.4201 $\rightarrow$ \textbf{0.6365} & 0.3135 $\rightarrow$ \textbf{0.2157} & 0.5655 $\rightarrow$ \textbf{0.1105} \\
Gemma3~\cite{Kamath2025Gemma3T} & 0.1971 $\rightarrow$ \textbf{0.5799} & 0.2213 $\rightarrow$ \textbf{0.6791} & 0.3253 $\rightarrow$ \textbf{0.2582} & 0.7589 $\rightarrow$ \textbf{0.0914} \\
\bottomrule
\end{tabular}%
}
\vspace{-1.0mm}
\end{table}

%\subsection{Real-Robot Evaluation}
%\label{sec:robot}

Finally, we conducted experiments on a UR3 robot with an NVIDIA RTX A2000 Ada G. (16 GB) GPU. CPOR-Grasp, UNOGrasp, and Gemini Robotics-ER-1.6 are evaluated on matched scenes using the same grasping backend and action budget. The 40-object set used to construct the real-robot evaluation scenarios is shown in Fig.~\ref{fig:robot-objects}. Following the UNOBench protocol, success requires correct blocker removal and target retrieval within the allotted budget. CPOR-Grasp leads at every tier and averages 77.8\% against 66.7\% and 62.2\% for the baselines. The margin widens with difficulty, and UNOGrasp's Hard-tier drop mirrors its Hard-tier decline on the offline benchmark (Table~\ref{tab:path-bench}).
\begin{figure}[!t]
\centering
\vspace{5mm}
\includegraphics[width=0.5\textwidth]{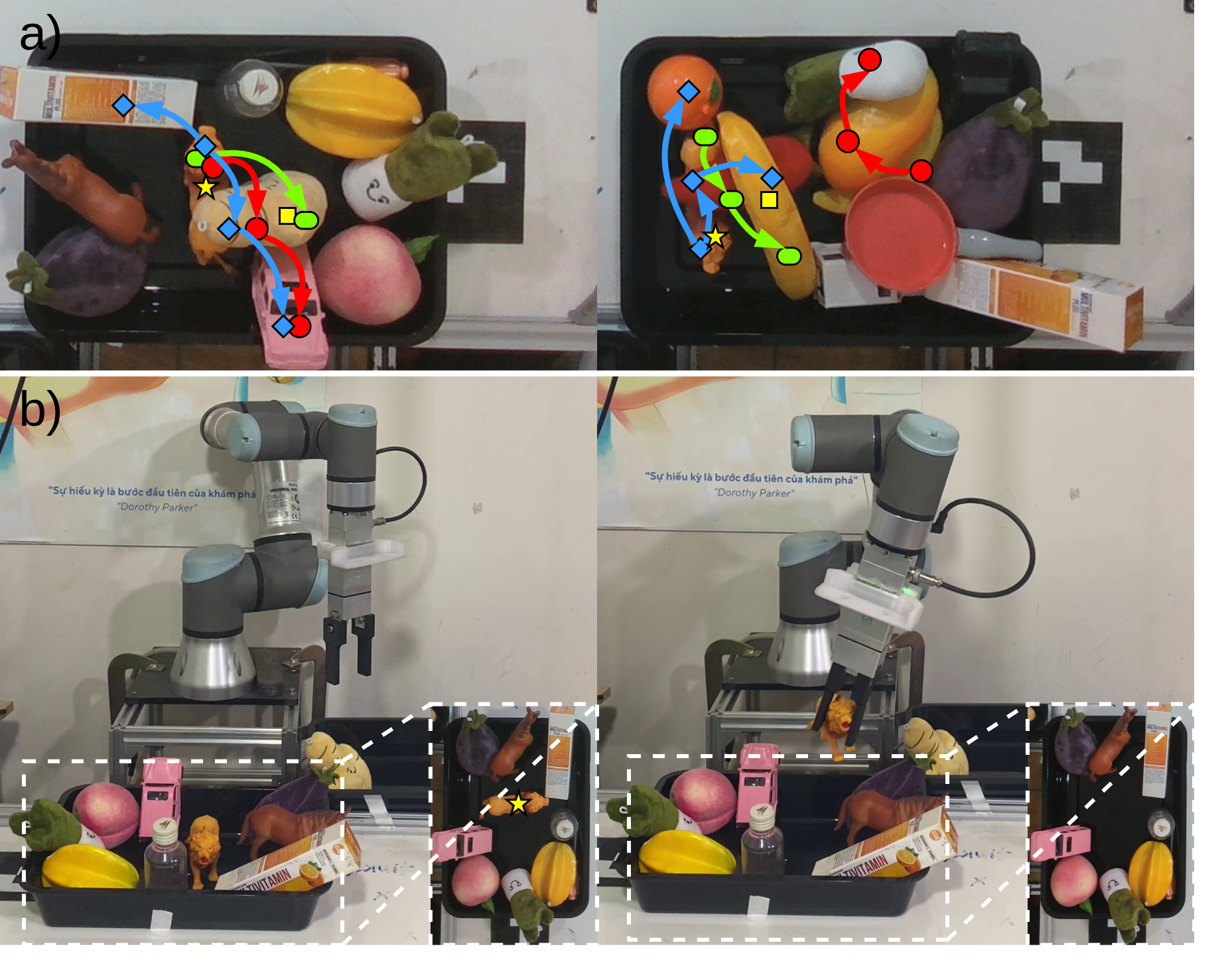}
\caption{
a) Qualitative obstruction reasoning on real-world scenes, including medium- and hard-difficulty cases. The target object is marked by \targetstar\ and the top obstructor by \topobstructor. Reasoning traces are shown for \unograspmarker\ UNOGrasp~\cite{jiao2026obstruction}, \bluediamond\ Gemini Robotics-ER 1.6, and \greenroundbox\ CPOR-Grasp. b) Real-robot deployment in a medium-difficulty scene.}
\label{fig:robot-quant}
\end{figure}

\begin{table}[!t]
\centering
\caption{Real-robot evaluation with the UR3 over 45 paired scenarios constructed from 40 objects.}
\label{tab:robot}
\vspace{-1.0mm}
\scriptsize 
\setlength{\tabcolsep}{0pt}
\renewcommand{\arraystretch}{1.0}

\begin{tabular*}{\columnwidth}{@{\extracolsep{\fill}}l c c c c@{}}
\toprule
\textbf{Method} & \textbf{Easy} & \textbf{Med.} & \textbf{Hard} & \textbf{Avg.} \\
\midrule
Gemini Robotics-ER-1.6~\cite{gemini_robotics_er_16}
& 86.7\% & 60.0\% & 53.3\% & 66.7\% \\

UNOGrasp~\cite{jiao2026obstruction}
& 86.7\% & 66.7\% & 33.3\% & 62.2\% \\

\textbf{CPOR-Grasp}
& \textbf{93.3\%} & \textbf{73.3\%} & \textbf{66.7\%} & \textbf{77.8\%} \\
\bottomrule
\end{tabular*}

\vspace{-1.5mm}
\end{table}

\section{Conclusion}
\textbf{CPOR-Grasp} closes the gap between relation scores and manipulation decisions through posterior inference over order-feasible obstruction graphs. It combines scene-conditioned calibration, reliability-gated RGB-D fusion, removal-order reasoning, and certified graph truncation. On UNOBench, it outperforms UNOGrasp on both synthetic and real scenes, reducing relation ECE from 0.1416 to 0.0185. Adaptive truncation matches exact inference on 99.74\% of decisions while evaluating only 22.68 graphs on average versus 1281.53 for exact enumeration. The framework improves performance across five VLM backbones and achieves a 77.8\% success rate on a UR3 robot, outperforming both baselines. Limitations remain: the certification applies only to inference under the model, the truncation bound can be loose in scenes with long directed cycles, and the geometry branch depends on a fine-tuned segmenter. Future work will learn candidate relations and extend certification to multi-step manipulation tasks.
\label{main:last}
\balance

\appendices
\section*{APPENDIX}
\subsection{Enumeration Constraints}
\label{app:ilp}
Let $z_{ij}\in\{0,1\}$ be the MILP variable corresponding to $e_{ij}$. The binary program adds a cycle cut for each violated directed cycle $C$ and a no-good cut after each returned mode $\widehat\bfe^{(h)}$:
\begin{equation*}
\begin{gathered}
\sum_{(a,b)\in C}z_{ab}\le|C|-1 \\
\sum_{(i,j):\widehat e^{(h)}_{ij}=1}(1-z_{ij})+\sum_{(i,j):\widehat e^{(h)}_{ij}=0}z_{ij}\ge1.
\end{gathered}
\end{equation*}
Each cycle cut is valid for every DAG, so it removes no DAG-feasible configuration. Once the current optimum is acyclic, it is optimal over the remaining DAG-feasible configurations allowed by the accumulated no-good cuts. Each no-good cut removes exactly one previously returned binary configuration. Repeated optimal solves therefore implement \eqref{eq:topk} in non-increasing weight. With a time limit or nonzero optimality gap, Theorem~\ref{thm:cert} still applies to any distinct feasible retained set, but exact Top-$K$ ordering is not guaranteed. Clipping with $0<\zeta<1/2$ keeps all modeled edge probabilities in $(0,1)$, so the empty graph has positive reference weight and $Z>0$. Configuration masses are accumulated in the log domain.

\subsection{Bayes Reference for Centered Log-Odds Fusion}
\label{app:reference}
Let $E^{vlm}_{ij}$ and $E^{cv}_{ij}$ denote the two source observations for a relation. Under $E^{vlm}_{ij}\perp E^{cv}_{ij}\mid e_{ij}$, a source-conditional assumption distinct from the across-edge independence in Assumption~1, Bayes' rule gives
\begin{equation}
\logit P(e_{ij}{=}1\mid E^{vlm}_{ij},E^{cv}_{ij})
=\logit\pi_0+\Lambda^{vlm}_{ij}+\Lambda^{cv}_{ij}
\label{eq:bayes-logit}
\end{equation}
, where $\Lambda^m_{ij}=\log P(E^m_{ij}\mid e_{ij}{=}1)-\log P(E^m_{ij}\mid e_{ij}{=}0)$. In this reference case, $\pi_0=P(e_{ij}{=}1)$, estimated in implementation by the fitting-split positive-edge rate. The log-likelihood ratio $\Lambda^m_{ij}$ equals the centered logit $h^m_{ij}$ only when the source score is the source-alone posterior under the same prior $\pi_0$. Calibration against empirical frequencies is weaker, and $p^{cv}_{ij}$ is not assumed to be a calibrated source posterior. For the geometric source, subtracting $\logit\pi_0$ is therefore reference centering of a learned feature, not a Bayesian prior correction. Equation~(\ref{eq:bayes-logit}) motivates only the additive form of \eqref{eq:fusion}. 
% The implementation does not evaluate $\Lambda^m_{ij}$.

\subsection{Proofs}
\label{app:proofs}
\noindent\textbf{Proof of Theorem~\ref{thm:cert}.}
For each unordered pair $\{i,j\}$, $\cR_2$ excludes only the reciprocal state $e_{ij}=e_{ji}=1$. These events involve disjoint Bernoulli coordinates across unordered pairs, which gives \eqref{eq:zbar}. Since every DAG avoids reciprocal two-cycles, $Z\le\overline Z$.
Let $\alpha=\PD(S_K\mid\Omega)=Z_K/Z$. Since $\PK=\PD(\cdot\mid S_K,\Omega)$, direct evaluation of the total-variation sum gives $\dTV(\PD,\PK)=1-\alpha=1-Z_K/Z$. The inequality in \eqref{eq:certificate} follows from $Z\le\overline Z$. For every $f:\cD\to[0,1]$, the variational characterization of total variation gives $|\mathbb E_{\PD}[f]-\mathbb E_{\PK}[f]|\le\dTV(\PD,\PK)$. Choosing $f=\chi_X$ or $f=\chi_o$ yields the marginal bounds. Along a nested retained sequence, $Z_K$ is non-decreasing while $\overline Z$ is fixed. \hfill$\square$

\noindent\textbf{Proof of Corollary~\ref{cor:stability}.}
% Theorem~\ref{thm:cert} gives $|s_a^\star-s_a|\le\epsilon_K$ for every executable action. Let $a^\dagger$ be the approximate winner. If $s_{(1)}-s_{(2)}>2\epsilon_K$, then for every $a\neq a^\dagger$, $s_{a^\dagger}^\star\ge s_{(1)}-\epsilon_K>s_{(2)}+\epsilon_K\ge s_a^\star$, so the exact winner is also $a^\dagger$. The exact maximum lies in $[s_{(1)}-\epsilon_K,s_{(1)}+\epsilon_K]$, which proves act/defer stability. Applying the same event bound to each $\chi_o$ proves blocker-membership stability. \hfill$\square$
Theorem~\ref{thm:cert} gives $|s_a^\star-s_a|\le\epsilon_K$ for every executable action $a$. (a) Let $a^\dagger$ be the approximate winner, so $s_{a^\dagger}=s_{(1)}$. Since $s_{(1)}>\tau+\epsilon_K\ge\tau$, the approximate rule acts and $\widehat a=a^\dagger$. Its exact score satisfies $s_{a^\dagger}^\star\ge s_{(1)}-\epsilon_K>\tau$, so the exact rule also acts. For every other executable $a$, $s_{a^\dagger}^\star\ge s_{(1)}-\epsilon_K>s_{(2)}+\epsilon_K\ge s_a+\epsilon_K\ge s_a^\star$, so $a^\dagger$ is the unique exact maximizer and $a^\star=a^\dagger$. When $|\cA_{\mathrm{exec}}|=1$, $s_{(2)}=-\infty$ and the rank condition holds trivially. (b) Every exact score satisfies $s_a^\star\le s_a+\epsilon_K\le s_{(1)}+\epsilon_K\le\tau$, so the exact rule defers, and $s_{(1)}\le\tau$ gives $\widehat a=\mathcal{D}$. (c) Choosing $f=\chi_o$ in Theorem~\ref{thm:cert} gives $|q_o-\widehat q_o^{(K)}|\le\epsilon_K$. If $\widehat q_o^{(K)}>\tau+\epsilon_K$ then $q_o>\tau$, and if $\widehat q_o^{(K)}<\tau-\epsilon_K$ then $q_o<\tau$, so $o\in\widehat F\Leftrightarrow o\in F^\star$. \hfill$\square$
\bibliographystyle{IEEEtran}
\bibliography{ref}  % .bib
\end{document}